\documentclass[english]{IEEEtran}
\usepackage[T1]{fontenc}
\usepackage[latin9]{inputenc}
\usepackage{float}
\usepackage{url}
\usepackage{amsmath}
\usepackage{graphicx}

\makeatletter

\providecommand{\tabularnewline}{\\}

\newcommand{\lyxaddress}[1]{
\par {\raggedright #1
\vspace{1.4em}
\noindent\par}
}

\usepackage[below]{placeins}
\usepackage{dblfloatfix}

\makeatother

\usepackage{babel}
\begin{document}

\title{Super-resolving multiresolution images with band-independant geometry
of multispectral pixels}

\author{Nicolas Brodu}
\maketitle

\lyxaddress{INRIA Bordeaux Sud-Ouest / Geostat, 200 avenue de la vieille tour,
33405 Talence, France}
\begin{abstract}
A new resolution enhancement method is presented for multispectral
and multi-resolution images, such as these provided by the Sentinel-2
satellites. Starting from the highest resolution bands, band-dependent
information (reflectance) is separated from information that is common
to all bands (geometry of scene elements). This model is then applied
to unmix low-resolution bands, preserving their reflectance, while
propagating band-independent information to preserve the sub-pixel
details. A reference implementation is provided, with an application
example for super-resolving Sentinel-2 data.
\end{abstract}

\section{Introduction}

\subsection{Context and state of art}

Earth Observation missions typically operate at medium to low resolution
ranges in order to favor both larger satellite swath and better temporal
revisit of the same site (e.g. 3-4 days over Europe for the Sentinel-2
Satellite series). For each acquisition, optical constraints furthermore
often restrict that only some spectral bands have maximal resolution.
For example, a common case is to compensate the smaller pixel size
of the higher spatial resolution bands (e.g. 10m/pixel) by capturing
light over a larger spectrum range (e.g. 4 large bands in the red,
green, blue and near infrared for Sentinel-2). The limit case being
a single high-resolution panchromatic band (Pleiades, Spot, Landsat...).
Narrower spectral bands, invaluable for specific measurements (e.g.
chlorophyll or water vapor absorbtion wavelengths) are then provided
at lower resolution (e.g. 20m/pixel and 60m/pixel for Sentinel-2).
Yet, this trade-off on spectral and spatial resolutions may become
a limiting factor for many Earth Observation applications, for example
for getting accurate land cover classification at the highest resolution
\cite{Radoux16}. Some techniques have thus been devised in order
to propagate the high-resolution spatial details to the lower-resolution
dedicated bands while preserving their spectral content. These can
be sorted along the following categories:

\textendash{} Probabilistic \cite{He09,Molina08}: The spectral information
in each sub-pixel of an original low-resolution pixel is determined
by maximizing a probabilistic model, constrained by the observed data
at all bands and resolutions (possibly including the panchromatic
band). A Bayesian formulation can be choosen to represent this constraint
which, provided this does not becomes intractable, allows hyperparameters
to be set according to prior knowledge.

\textendash{} Sensor-based: If the sensor has a known point spread
function (PSF), then deconvoluting it enhances the resolution of the
acquired images \cite{Nakazawa14}. But the PSF for many satellites
can only be estimated empirically (Sentinel-2, Spot-5, Landsat-8 \cite{Radoux16}).
When that is the case, limits on sub-pixel detection can be established
\cite{Radoux16}.

\textendash{} Learning-based: These methods exploit local patterns
in the low-resolution images, and propagate these features (e.g. edges)
to infer the higher resolution image \cite{Baker02}. Many models
may be used to ``learn'' the features: neural network \cite{Liebel16},
example-based \cite{Freeman02} with kernel ridge regression \cite{Kim08},
deep learning \cite{Dong16} and more references therein, including
for cross-image learning. These methods can be applied for single
image resolution enhancement, possibly with different channels (typically
red, green, blue \cite{Dong16}). Filling details from learned (or
duplicated) texture features might be very good to produce visually
plausible results \cite{Freeman02}. Their main problem, similar to
in-painting with image-based examples \cite{Barnes09}, is that ``hallucinated''
\cite{Baker02} details do not necessarily correspond to true higher-resolution
objects (esp. with non-local or cross-image features) and then become
misleading pixels for land monitoring purposes.

\textendash{} Scaling laws: Instead of learning local patterns, this
method learns multi-scale relationships in the data such as local
power laws between spatial extent and band values \cite{Turiel08}.
Such scaling laws are inferred from data above the aquisition resolution
but, assuming the same laws remain valid below that resolution, these
can then be used to infer sub-pixels of the original image. Very good
results have been obtained with such methods for turbulent oceanic
data \cite{Sudre15}, where energy cascades translates to power laws
spanning multiple decades. However, for land monitoring purposes,
usually no such physical interpretation can be found: for example
a mixture of trees, houses and roads in a peri-urban environment is
not locally scale-invariant. 

\textendash{} Frequency representations: Working in the frequency
domain, whether with Fourier methods or using wavelet decompositions.
An idea is to upsample the low-frequencies (e.g. with a bicubic filter)
and preserve the high frequency components. Unfortunately, natural
images are not statistically consistent: knowledge that there is a
tree (or road, house...) a few hundred meters away (i.e. the wavelet
support size) does not help subdivide a local pixel into its higher-resolution
components.

\textendash{} Panchromatic sharpening \cite{Nikolakopoulos08}: Using
a very wide band with high-resolution in order to compensate the lower
resolution of narrow bands. Multiple variants exist, from a simple
renormalization of the multispectral bands \cite{Gillespie87} to
more advanced unmixing techniques which estimate the contribution
of each spectral band to the panchromatic one \cite{Padwick10}, possibly
with pre-processing steps designed to uncorrelate each component \cite{Chavez91,Nikolakopoulos08},
or using angular spaces such as the hyperspherical color space \cite{Wu15}.

The advantages of panchromatic sharpening are its efficiency, and
its applicability even when only a single high-resolution band is
acquired. Many observation satellites thus include a panchromatic
band, but some do not (e.g. ESA's Sentinel-2 series, specifications
given in Appendix). In the absence of such a band, and given the inadequacies
of the other methods presented above, another solution is needed.

\subsection{\label{subsec:Superres_without_pan}Super-resolution of multispectral
images in the absence of a panchromatic band}

In a multispectral measurement system, each pixel in a band $B$ captures
the light intensity over some part of the spectrum, according to some
sensor sensitivity distribution $s_{B}(\lambda)$ for each wavelength
$\lambda$: $B(x,y)=\int s_{B}(\lambda)I(\lambda,x,y)d\lambda$. The
light intensity $I$ is reflected at wavelength $\lambda$ by the
pixel surface between $x+r_{B}$ and $y+r_{B}$, where $r_{B}$ is
the square pixel resolution of band $B$. Larger pixels thus collect
more light and may be necessary for some bands with narrow $s_{B}$
spectral support. This very simplified model ignores many sources
of optical distortions and the post-processing from satellite geometry
to square pixels, but these effects are irrelevant in the context
of this section.

When a panchromatic band $P$ is available, light is collected over
a wide spectrum support : $s_{P}(\lambda)\neq0$ for a large range
of wavelengths $[\lambda_{Pmin},\lambda_{Pmax}]$, usually covering
all other bands $\lambda_{Pmin}\leq\lambda_{Bmin}\leq\lambda_{Bmax}\leq\lambda_{Pmax}$.
Collecting more light spectrally allows to reduce the pixel size $r_{P}<r_{B}$
while still maintaining a minimal intensity to trigger the panchromatic
sensor. Each band pixel $B(x,y)$ thus covers an area with multiple
panchromatic pixels $P(x+i*r_{P},\,\,y+j*r_{p})$ for $i,j$ indices
depending on the band resolution $r_{B}$. The total light reaching
the sensor for $B(x,y)$ may thus be related to the values of each
sub-pixels $P(x+i*r_{P},\,\,y+j*r_{p})$ by a function involving both
spatial and spectral components. Panchromatic sharpening methods attempt
to unmix this total light contribution by defining $B(x,y)=f(H(x+i*r_{p},\,\,y+i*r_{p}))$,
where $H$ is a high-resolution version of band $B$. $f$ is a coarse-graining
function, for example $f=\sum_{i,j}$ in the simplest case. More elaborated
functions may be necessary, for example when combining both panchromatic
sharpening and atmospheric corrections. The relations between $s_{P}(\lambda)$
and $s_{B}(\lambda)$ are used in order to further constrain $H$
with values of $P$ at the same pixel locations.

Unfortunately, for Sentinel-2 images, there is no panchromatic band.
Methods using such bands, cited in the previous Section, are thus
not directly useable. An idea would be to create a virtual $P$ band
by combining the four high-resolution bands $B^{10}$ at 10m/pixel,
and then apply pansharpening methods for bands at 20m/pixel and 60m/pixel.
But combining blue, green, red, and infrared bands into a virtual
intensity would result in a virtual $s_{P}(\lambda)$ that only covers
the original $s_{B^{10}}(\lambda)$, and thus prone to spectral artifacts
for super-resolving the values of the other bands $B^{20}$ and $B^{60}$.
Similarly, spatial details in that hypothetical $P$ band would depend
on how that virtual $P$ is constructed from other $B^{10}$ bands
(in particular, how infrared details are fused with visible light
details).

In the absence of a real panchromatic band, a better idea is to design
a new method that:

\textendash{} Explicitely encodes geometric details from available
high-resolution bands, as pixel properties independent from their
reflectance;

\textendash{} Preserves the spectral content of each low-resolution
band independently from the geometry.

The next sections introduce one method for acheiving this, with direct
applicability to Sentinel-2 images. The method presented below works
with only local information, hence it is not subject to the non-local
effects mentionned in the previous section. The method relies on the
observation that the proportion of objects of the same nature within
a pixel area (e.g. 30\% urban area, 70\% trees), is a physical property
of that pixel and therefore independent of the spectral band. Only
the reflectance of these objects may change from band to band. Moreover,
there is no reason why pixel boundaries would match natural object
boundaries. The method thus identifies generic ``shared'' information
between adjacent pixels, then commonalize the geometric aspects of
these shared values across bands. High resolution bands are used to
separate band-independent information from band-dependant reflectance.
The geometric information is then used to unmix the low-resolution
pixels, while preserving their overall reflectance.

Section \ref{sec:Problem-description} presents the super-resolution
problem and Section \ref{sec:Solving-the-super-resolution} details
how that problem is addressed by the model introduced in this paper.
Section \ref{sec:Performance-assessement} indicates how to quantify
the results quality, and section \ref{sec:Results} shows super-resolution
results for three different types of regions of interest (coastal,
urban, agricultrural). Results are followed by a discussion in Section
\ref{sec:Discussion}, which also demonstrates the influence of each
step of the algorithm.

\section{\label{sec:Problem-description}Problem description}

\subsection{Super-resolution formulation}

\begin{figure}[b]
\includegraphics[width=1\columnwidth]{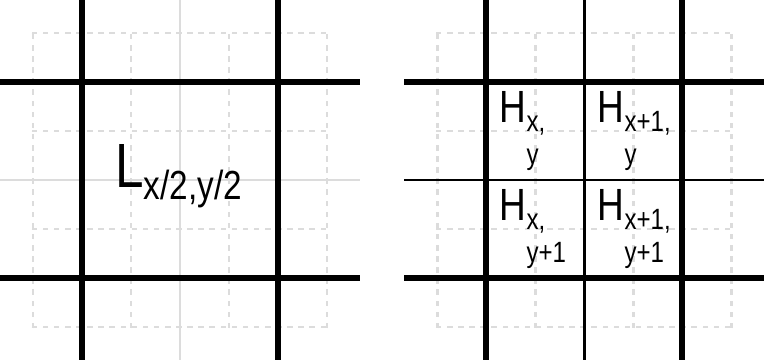}

\caption{\label{fig:lh}Introducing the indexing and the relations between
the low and high resolution pixels, for the simple case of doubling
the resolution. The grayed lines indicate boundaries from Fig.~\ref{fig:sw},
for ease of interpretation.}
\end{figure}

Let $L$ be an observed low resolution image with $N_{x}/2$ columns
and $N_{y}/2$ rows. We consider the problem of finding a high resolution
image $H$ with $N_{x}$ columns and $N_{y}$ rows. Each low resolution
pixel $L_{x/2,y/2}$ thus corresponds to $4$ high resolution pixels,
as depicted in Fig.~\ref{fig:lh}. Averaging these pixels should
give the original observed low-resolution pixel back:

\begin{equation}
L_{x/2,y/2}=\frac{1}{4}\left(H_{x,y}+H_{x+1,y}+H_{x,y+1}+H_{x+1,y+1}\right)\label{eq:L_eq_Havg}
\end{equation}

Images remotely sensed from satellites are subject to multiple transforms
(including atmospheric corrections \cite{Sen2Cor221}) before being
released as a useable product. These transforms are out of scope of
the present document but may introduce correlations between high-resolution
pixels (e.g. due to scattering), hence should be applied before (or
integrated to) super-resolution. Similarly, known PSF \cite{Nakazawa14}
should be deconvoluted in addition to the method presented below.
In any case, Eq.~\ref{eq:L_eq_Havg} ensures that down-sampling by
averaging the high-resolution solution will recover the observations.

Eq.~\ref{eq:L_eq_Havg} is undetermined, with 3 free parameters per
low resolution pixel. Some extra constraints are needed, which are
extracted from available high-resolution data bands.

\subsection{Shared information between neighbor pixels}

\begin{figure}[b]
\includegraphics[width=1\columnwidth]{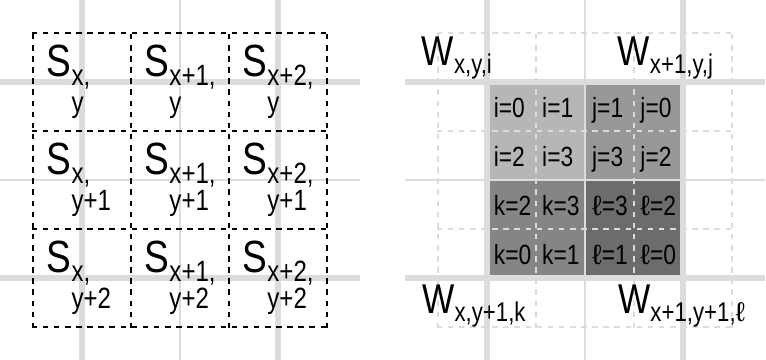}

\caption{\label{fig:sw}Values $S$ shared between neighbor pixels, and how
these are combined by weights $W$ to form the high-resolution pixels.
Compare with the boundaries from Fig.~\ref{fig:lh}, indicated as
grayed lines for ease of interpretation. Shared values span over multiple
pixels by definition. These values represent the reflectance of elements
that are common to the spanned pixels. Weights are internal to each
pixel and represent the proportion of these elements within the pixel.
Weights are thus independant of the spectral band.}
\end{figure}

Natural objects do not fall exactly on pixel boundaries. Therefore,
some content is shared between nearby pixel values. This shared information
is explicitly defined as in Fig.~\ref{fig:sw}, left. For example,
$S_{x+1,y+1}$ is the reflectance corresponding to the shared part
between high-resolution pixels $H_{x,y}$, $H_{x+1,y}$, $H_{x,y+1}$
and $H_{x+1,y+1}$ (compare Fig.~\ref{fig:lh} and Fig.~\ref{fig:sw}).
This particular element is fully within the observed low-resolution
pixel $L_{x/2,y/2}$. Other shared values may span multiple low-resolution
pixels. With this notation, there are $\left(N_{x}+1\right)\times\left(N_{y}+1\right)$
spatially shared values $S$. These located at the image boundaries,
or which would span invalid pixels such as in the case of sensor failure,
are simply not commonalized and effectively remain internal to the
valid pixels. All shared values are expressed in reflectance units
and constrained to the range of their respective band.

In remote sensing, the reflectance of each pixel is often considered
to be a linear mix \cite{Settle93} of the reflectances of its constituents
(e.g. a mix of vegetation and soil). Assuming the shared values correspond
to unknown constituents spanning over pixel boundaries, and using
this linear mixing model, the proportion of each shared value that
is present in each pixel is thus determined by weights that are specific
to that pixel (Fig.~\ref{fig:sw}, right). This leads to the following
mixing equation for the shared values and the weights:

\begin{align}
H_{x,y}= & W_{x,y,0}S_{x,y}+W_{x,y,1}S_{x+1,y}\label{eq:H_mixing_SW}\\
+ & W_{x,y,2}S_{x,y+1}+W_{x,y,3}S_{x+1,y+1}\nonumber 
\end{align}

With the following constraints:

\begin{equation}
\sum_{k=0}^{3}W_{x,y,k}=1\label{eq:SumW}
\end{equation}

\begin{equation}
W_{x,y,k}\ge0\,\,\,\forall k\in[0\ldots3]\label{eq:Wbounds}
\end{equation}

\section{\label{sec:Solving-the-super-resolution}Solving the super-resolution
problem}

\subsection{\label{subsec:Sopt_W}Separating band-specific information from information
common to all bands}

The proportion of mixed elements within a pixel (e.g. 20\% road /
80\% vegetation) is a physical property of that pixel, but the reflectance
of each element depends on the spectral band at which it is observed.
Therefore, the weights are common to all bands, while shared values
are band-dependant. Weights encode the geometric consistency of pixels
accross bands. Shared values encode the spatial consistency of nearby
pixels. The high-resolution data are used to fit the full mixing model,
containing both weights and shared values. This step is presented
below. The next section addresses how to un-mix low-resolution bands
in order to produce the super-resolution result, reusing the weights
fit from the high-resolution bands.

Starting from an observed high-resolution band $H^{o}$, a down-sampled
version $L^{d}$ of the data is created with Eq.~\ref{eq:L_eq_Havg}.
The best mixing model is estimated by minimizing the difference between:
a) the observed pixel values $H^{o}$, and b) the resolution-enhanced
values $H^{r}$ computed from the down-sampled data $L^{d}$. Let
us subscript data specific to each band with an additional index $\beta$.
Thus, $L^{d}$, $H^{o}$, $H^{r}$ and $S$ are subscripted, but not
the weights $W$. Solving this first problem is a constrained minimization,
for $k=0\ldots3$, and $\beta\in\mathcal{H}$ the set of high-resolution
bands:

\begin{equation}
\left\{ S^{opt},W^{opt}\right\} =\textrm{argmin}\sum_{\beta\in\mathcal{H}}\sum_{x,y}\left\Vert H_{\beta,x,y}^{o}-H_{\beta,x,y}^{r}\right\Vert ^{2}\label{eq:solve_for_weights}
\end{equation}

with each $H_{\beta,x,y}^{r}$ term given by Eq.~\ref{eq:H_mixing_SW}.
An iterative solver \cite{CeresSolver} is used for constrained least
squared error optimization\footnote{The Ceres solver \cite{CeresSolver} can be fine-tuned with many internal
parameters. Extensive testing determined that conjugate gradients
with a block Jacobi preconditionner give the best quality/processing
time tradeoff for the super-resolution problem presented in this article.
These are set by default in the reference implementation.}, allowing Eq.~\ref{eq:SumW} to be enforced by a reparametrization
and Eq.~\ref{eq:Wbounds} by soft boundaries (a reference implementation
is provided, link given in appendix). Initial weights for the iterative
algorithm are set to $\frac{1}{4}$ (i.e. equal influence to all shared
values, see Fig.~\ref{fig:sw}, left). The initial shared values
$S_{\beta}^{ini}$ are computed by averaging each high-resolution
pixel $H_{\beta}^{o}$ that partially covers $S_{\beta}^{ini}$ in
Fig.~\ref{fig:sw}. For example, $S_{\beta,x+1,y+1}^{ini}=\frac{1}{4}\left(H_{\beta,x,y}^{o}+H_{\beta,x+1,y}^{o}+H_{\beta,x,y+1}^{o}+H_{\beta,x+1,y+1}^{o}\right)$.

At the end of this step, both shared values $S_{\beta}^{opt}$ between
high-resolution pixels, and weights $W^{opt}$ common to all bands,
are available.

\subsection{\label{subsec:Estimating-shared-values}Estimating shared values
from low-resolution data}

\begin{figure}[H]
\includegraphics[width=1\columnwidth]{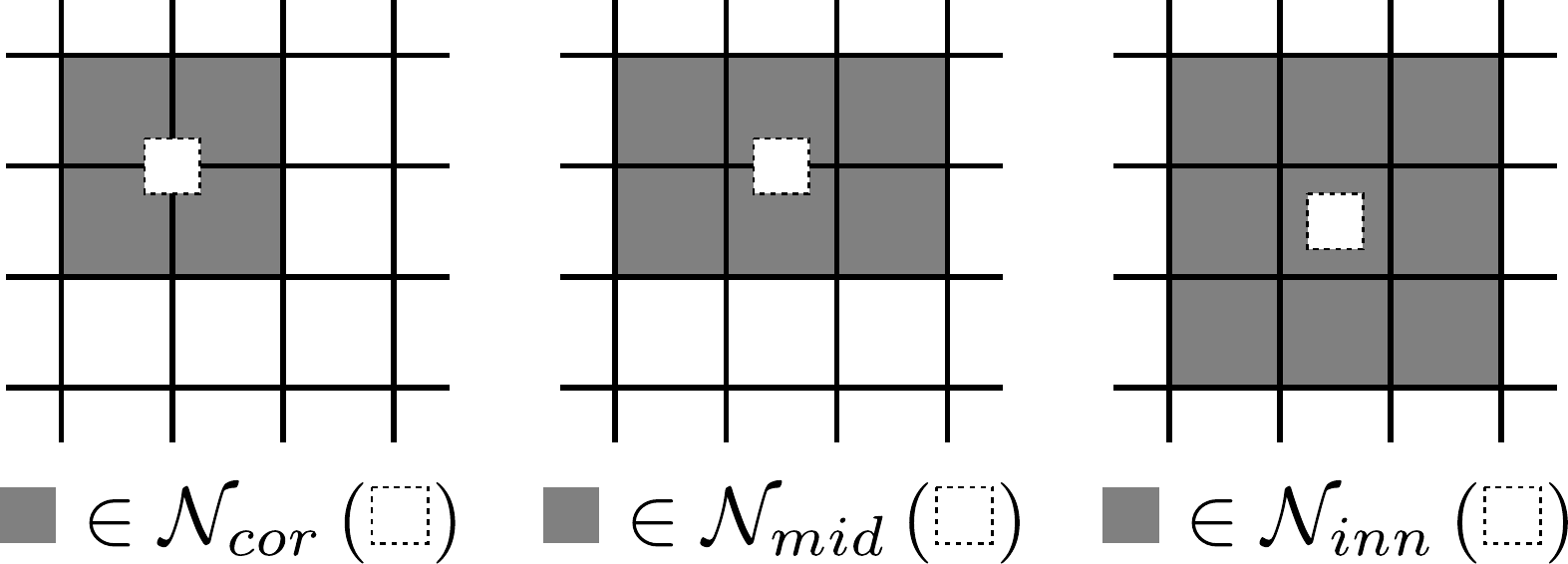}

\caption{\label{fig:Neighborhoods-for-V}Low-resolution neighborhoods for high-resolution
shared values. Depending on the location of the shared value with
respect to the center reference pixel (corner, middle, inner), the
neighborhood consists of either 4, 6 or 9 low-resolution pixel locations.
Other corner and middle locations are deduced by a rotation of the
pattern.}
\end{figure}

Shared values $S^{opt}$ are found by optimization on high-resolution
data, so they cannot be estimated directly on the low-resolution bands
with the above procedure. Instead, the relation between $S^{opt}$
and nearby low-resolution pixels can be learned from downsampled high-resolution
bands $L^{d}$. That relation is also expressed as a geometric property
common to all bands, so it can be used in order to produce a first
estimate $S^{fit}$ for the low-resolution bands. More specifically,
a second set of mixing coefficients $V$ is built in order to fit
$S_{\beta,x,y}^{opt}$ from low-resolution pixels $L_{\beta,n}^{d}$
at nearby locations $n\in\mathcal{N}(x,y)$. See Fig.~\ref{fig:Neighborhoods-for-V},
with $\mathcal{N}$ being either the corner, middle or inner variant
depending on the position of $(x,y)$ with respect to the low-resolution
reference pixel. 
These neighborhoods hopefully capture local features (e.g. edges),
in the form of up to $9$ coefficents $v_{x,y,n}$ for each high-resolution
pixel $(x,y)$ in the image:

\begin{equation}
V=\textrm{argmin}\sum_{\beta\in\mathcal{H}}\sum_{x,y}\left\Vert S_{\beta,x,y}^{opt}-\sum_{n\in\mathcal{N}\left(x,y\right)}v_{x,y,n}L_{\beta,n}^{d}\right\Vert ^{2}\label{eq:fitting_V_from_LR}
\end{equation}

With this global optimization, the set of coefficients $V$ encodes
how the shared values are related to their low-resolution neighborhood,
independently of the spectral band. They are fit from the high-resolution
bands, and then propagated to the low-resolution bands $b\in\mathcal{L}$
in order to provide a first estimate $S^{fit}$ for the shared values
in each band $b$:

\begin{equation}
S_{b,x,y}^{fit}=\sum_{n\in\mathcal{N}\left(x,y\right)}v_{x,y,n}L_{b,n}\label{eq:Sfit}
\end{equation}

The fit from Eq.~\ref{eq:fitting_V_from_LR} is rarely perfect and
values $S_{\beta,x,y}^{fit}=\sum_{n\in\mathcal{N}\left(x,y\right)}v_{n}L_{\beta,n}^{d}$
can also be computed for the original high-resolution bands. The ratios
$q_{\beta,x,y}=S_{\beta,x,y}^{opt}/S_{\beta,x,y}^{fit}$ are then
exploited in order to mimick the panchromatic sharpening method in
\cite{Gillespie87}, but using the multiple high-resolution bands
instead. For Sentinel-2, no panchromatic band is available to encompass
the spectrum of all low-resolution bands. An idea is to empirically
replace the panchromatic band by a combination of high-resolution
bands. Problems mentionned in Section \ref{subsec:Superres_without_pan}
are adressed by weighting bands that yield close spectral responses
for the shared values. For each low resolution band $b\in\mathcal{L}$,
and for each shared value, a normalized proximity measure is defined
as $p_{b,\beta,x,y}=|S_{b,x,y}^{fit}-S_{\beta,x,y}^{fit}|/\max_{\alpha}|S_{b,x,y}^{fit}-S_{\alpha,x,y}^{fit}|$,
where the normalization is performed by using the maximum discrepancy
over all high resolution bands $\alpha\in\mathcal{H}$. Combining
the high-resolution sharpening ratios is then performed by geometric
averaging, using these proximity measures as weighting factors: 
\begin{equation}
\bar{q}_{b,x,y}=\exp\left(\left(\sum_{\beta}p_{b,\beta,x,y}\log q_{\beta,x,y}\right)/\sum_{\beta}p_{b,\beta,x,y}\right)\label{eq:average_ratio}
\end{equation}

This overall average sharpening ratio is used as a prefactor for setting
corrected shared values $S_{b,x,y}^{cor}=\bar{q}_{b,x,y}S_{b,x,y}^{fit}$
for the low resolution bands. Having now estimated high-resolution
shared values for the low-resolution bands, it is a simple matter
of combining these $S_{b,x,y}^{cor}$ with the weights $W$ by applying
Eq.~\ref{eq:H_mixing_SW}, in order to produce super-resolved pixels.
A final rescaling of these $H_{b,x,y}^{r}$ is performed so as to
ensure reflectance is preserved (Eq.~\ref{eq:L_eq_Havg}).

\subsection{Super-resolving 60m/pixel bands}

\begin{figure*}[t]
\begin{centering}
\includegraphics[width=0.75\textwidth]{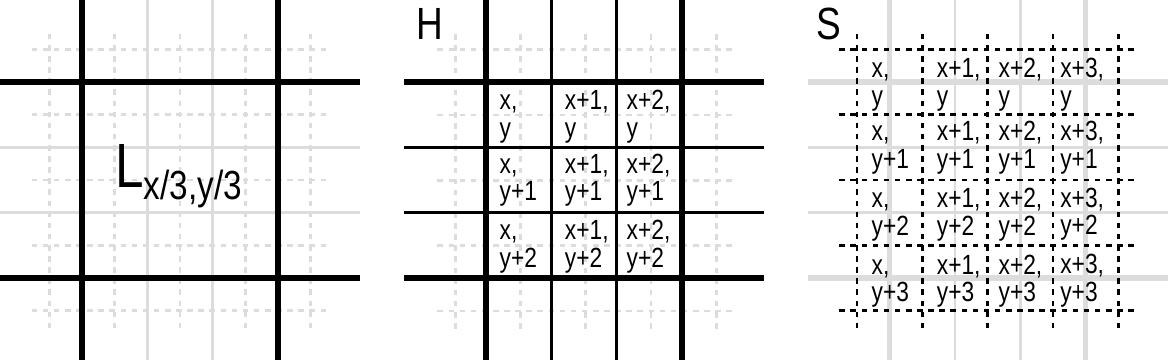}
\par\end{centering}
\caption{\label{fig:lhs_x3}Introducing the indexing and the relations between
the low and high resolution pixels, for the case of tripling the resolution.
The shared values between pixels are also indicated. Weights are still
internal to each high-resolution pixel with exactly the same structure
as in Fig.~\ref{fig:sw}, right.}
\end{figure*}

In this setup, each low-resolution pixel corresponds to 36 values
at 10m/pixel. Solving this directly is not tractable, but an indirect
solution with an intermediate step is feasible:

\textendash{} In a first pass, a 60m/pixel band is super-resolved
to 20m/pixel. There are then 9 sub-pixels to infer for each low-resolution
pixel, with 8 free parameters (Fig.~\ref{fig:lhs_x3} and Eq.~\ref{eq:L_eq_Havg_x9}
below). However, there are also 10 bands at 20m/pixel (including the
4 bands at 10m/pixel, downsampled). These provide enough constraints
for the inference of weight values, common to all these bands, computed
with a modified version of the method presented below.

\textendash{} In a second pass, the 20m/pixel solution from the first
pass is super-resolved down to 10m/pixel with the method described
in the previous sections.

Adapting the above notation to the 9 sub-pixels problem, in this section
the low resolution bands are $b\in\mathcal{L}$ at 60m/pixel, while
the high resolution bands $\beta\in\mathcal{H}$ consist of the 20m/pixel
bands (either original or down-sampled from 10m/pixel). Preserving
the reflectance imposes (Fig.~\ref{fig:lhs_x3}):

\begin{equation}
L_{b,x/3,y/3}=\frac{1}{9}\sum_{i,j=0}^{2}H_{b,x+i,y+j}^{r}\label{eq:L_eq_Havg_x9}
\end{equation}

$H_{b}^{r}$ is the intermediate super-resolution solution at 20m/pixel
for band $b$. Fig.~\ref{fig:lhs_x3} also shows the relation between
$L_{b}$, the high-resolution pixels $H_{b}^{r}$ and the shared values.
With this convention, the weights $W$ follow exactly the same $(x,y)$
pattern with respect to $S$ and $H$ as in Fig.~\ref{fig:sw}, right.
The first step of the method, the estimation of both $W$ and $S$
with all available high-resolution bands, is thus also the same as
above. All 10 bands $\beta\in\mathcal{H}$ are used as constraints
for Eq.~\ref{eq:solve_for_weights}.

A difference lies in estimating $V$ from nearby pixels. There are
still four neighborhood patterns of the ``corner'' type (see Fig.~\ref{fig:Neighborhoods-for-V}),
for shared values$S_{x,y}$, $S_{x+3,y}$, $S_{x,y+3}$ and $S_{x+3,y+3}$.
But there are now two ``middle'' neighborhood patterns for each
side of the lower-resolution pixel (e.g. $S_{x+1,y}$, $S_{x+2,y}$),
as well as four ``inner'' neighborhood patterns instead of one (see
Fig.~\ref{fig:lhs_x3}, right). With these definitions, Eq.~\ref{eq:fitting_V_from_LR}
is solved as before. Averaging the sharpening ratios now involves
all 10 bands $\beta\in\mathcal{H}$ (instead of 4 in the previous
section), but this does not change Eq.~\ref{eq:average_ratio}. Thus,
solving the 60m$\rightarrow$20m super-resolution problem is performed
with very little adaptation.

Once computed, the intermediate solutions at 20m/pixel are further
processed by a final 20m$\rightarrow$10m super-resolution step, as
described in the above sections.

\section{\label{sec:Performance-assessement}Performance assessement}

Quantitative measures are needed in order to evaluate the quality
of the super-resolution. Typical quantifiers \cite{Wald00,Alparone08}
include :

\textendash{} The quality index $Q(x,y)=\frac{4\cdot cov(x,y)\cdot mean(x)\cdot mean(y)}{\left(var(x)+var(y)\right)\left(mean(x)^{2}+mean(y)^{2}\right)}$
between an image $x$ and an image $y$.

\textendash{} The normalized mean squared error $ERGAS(x,y)=100R\sqrt{\frac{1}{N}\sum_{i=1}^{N}\frac{MSE(x_{i},y_{i})}{mean(x_{i})}}$,
where $N$ is the number of bands, $x$ is the reference image, and
$y$ is the image to be tested.

\textendash{} The spectral angle $SAM=\textrm{arccos}\left(\frac{x\cdot y}{\left\Vert x\right\Vert \left\Vert y\right\Vert }\right)$,
considering the $x$ and $y$ images as vectors. $SAM$ is given in
degrees in the next Section.

\textendash{} When a panchromatic band $P$ is available, the quality
with no reference $QNR=\left(1-D_{\lambda}\right)^{\alpha}\left(1-D_{s}\right)^{\beta}$
can be used, where $D_{\lambda}$ is a spectral distortion index and
$D_{s}$ is a spatial distortion index. But $D_{s}$ depends on $P$
so, in the present case with no panchromatic band available, $QNR$
is not useable.

In the following sections, $Q$, $ERGAS$ and $SAM$ are computed.
The typical methodology for using these quantifiers is to down-sample
an image, super-resolve this down-sampled image back to the original
resolution, and compare with the original data. Typically, the highest-resolution
bands are used for downsampling/super-resolution. But, due to the
way super-resolution is performed in this paper, downsampling 10m
bands to 20m and super-resolving them back for comparison is not acceptable.
Indeed, that very step is already included as part of the optimization
in Eq.~\ref{eq:solve_for_weights}. Moreover, shared details coming
from all original 10m bands are taken into account in Eqs.~\ref{eq:fitting_V_from_LR}-\ref{eq:average_ratio}.
Hence, a test that uses these same 10m bands as a basis for comparison
wouldn't be fair. The method is thus adapted as follows :

\textendash{} The 20m bands are downsampled to 40m and the four 10m
bands are downsampled to 20m.

\textendash{} Each downsampled 40m band is super-resolved back to
20m using the four original 10m bands that were downsampled at 20m.

\textendash{} Each original 20m band $x$ is compared with its downsampled/super-resolved
version $y$

This way, the original 10m bands are only used in order to build $y$,
but are not themselves the basis of comparison $x$. The downside
is that this method does not quantify directly the quality of final
product (the super-resolution of all bands to 10m), but a proxy for
it.

Statistics are given for each 20m band, then globally averaged over
all bands : geometric average for $Q$, using the given formula for
$ERGAS$, and arithmetic angle average for $SAM$.

\section{\label{sec:Results}Results}

Three use cases were selected for testing the algorithm:

\textendash{} A coastal environment, the delta of the Eyre river (France);

\textendash{} A urban area, the city of Bordeaux;

\textendash{} Fields, in the Bordeaux peri-urban area.

All these regions of interest are tested on an image acquired by the
Sentinel 2A satellite on 2016/08/22 and processed with the ``sen2cor''
atmospheric correction utility \cite{Sen2Cor221}. Figs.~\ref{fig:Coastal_area_RGB_B8_B1_B5},\ref{fig:Urban_area_RGB_B8_B1_B5}
and \ref{fig:Fields_area_RGB_B8_B1_B5} show the three selected regions
as a composite images from the 10m/pixel visible bands, where each
blue, green, red component was scaled between 1\% and 99\% of the
original reflectance.

As an indication of computational performance, processing all the
bands in either of these 408x300 pixel areas takes on average 1min17s
on a machine with twelve 1.9GHz cores.

\subsection{Coastal Area}

\begin{figure*}[b]
\includegraphics[width=0.5\textwidth]{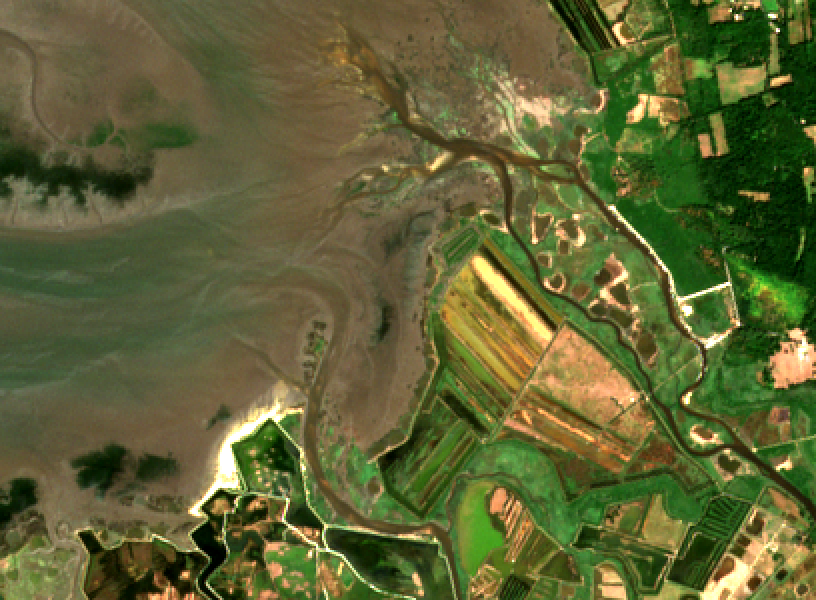}\includegraphics[width=0.5\textwidth]{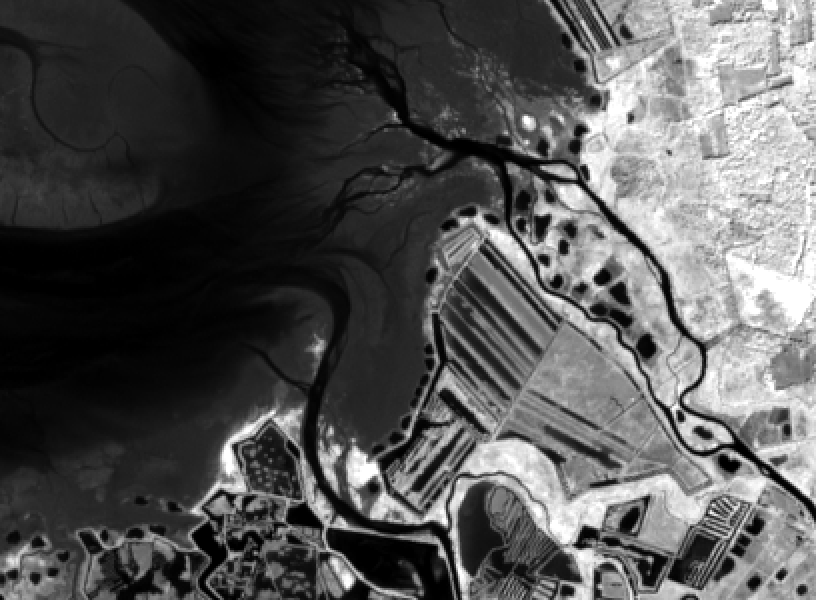}

\includegraphics[width=0.5\textwidth]{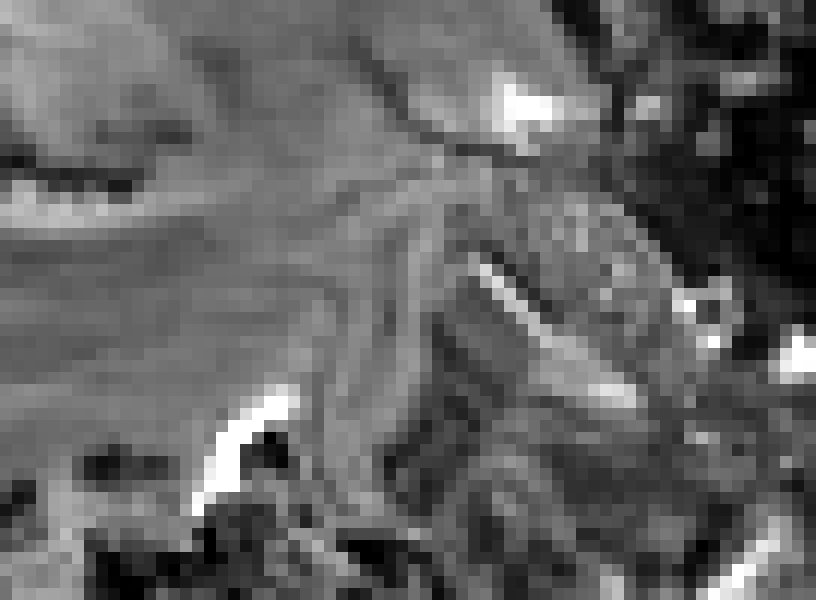}\includegraphics[width=0.5\textwidth]{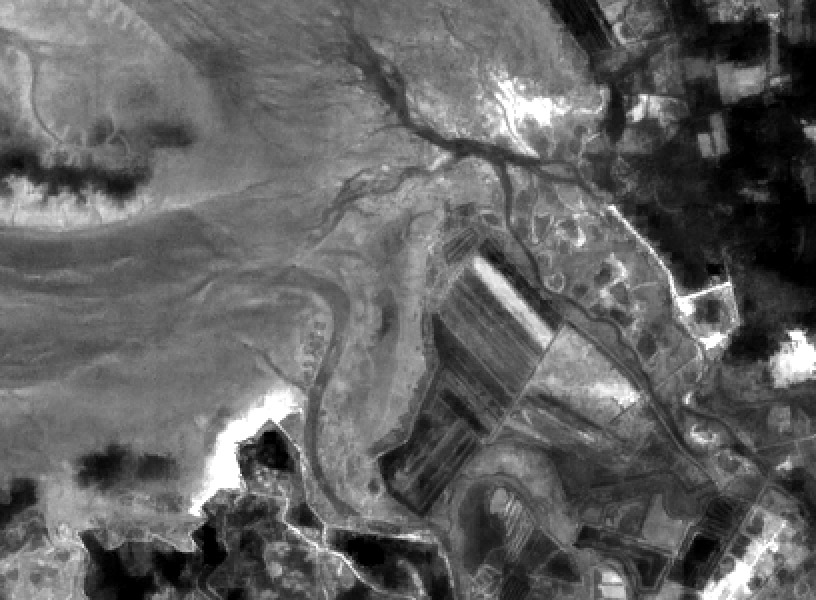}

\includegraphics[width=0.5\textwidth]{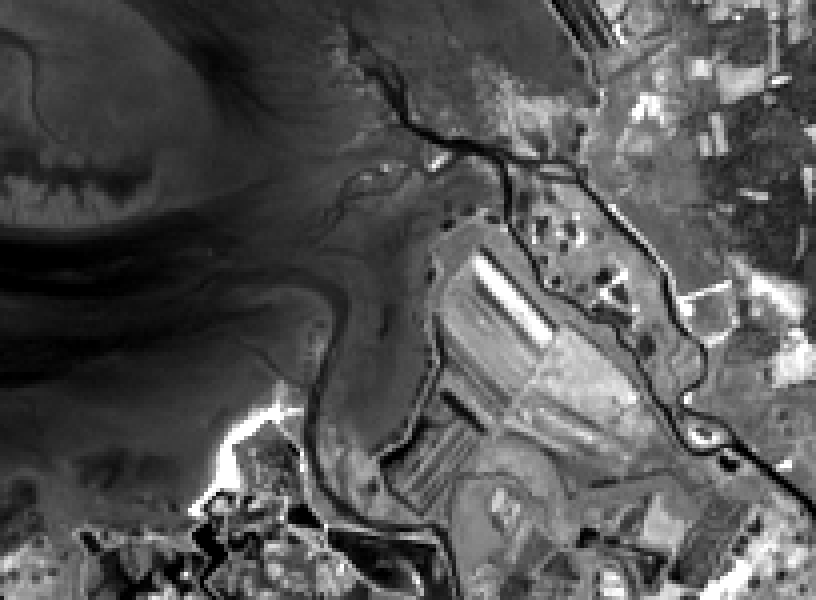}\includegraphics[width=0.5\textwidth]{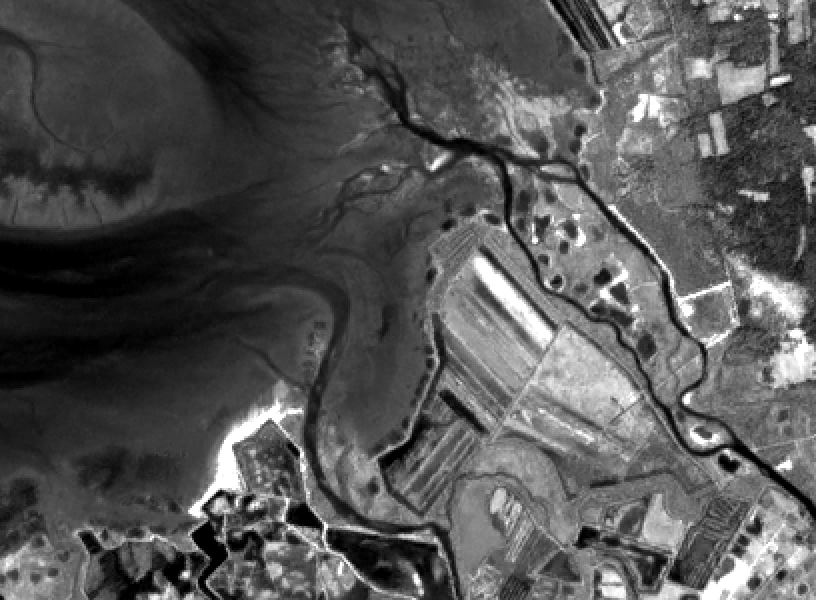}\caption{\label{fig:Coastal_area_RGB_B8_B1_B5}Top-Left: Composite image of
the coastal area (delta of the Leyre river). Top-Right: Original infrared
band 8 (842 nm, 10m/pixel). Middle-Left: Original band 1 (violet,
443nm, 60m/pixel). Middle-Left: Super-resolved band 1 at 10m/pixel.
Bottom-left: Original band 5 (red-edge, 705nm, 20m/pixel). Bottom-right:
Super-resolved band 5 at 10m/pixel.}
\end{figure*}

\begin{figure*}[b]
\includegraphics[width=0.5\textwidth]{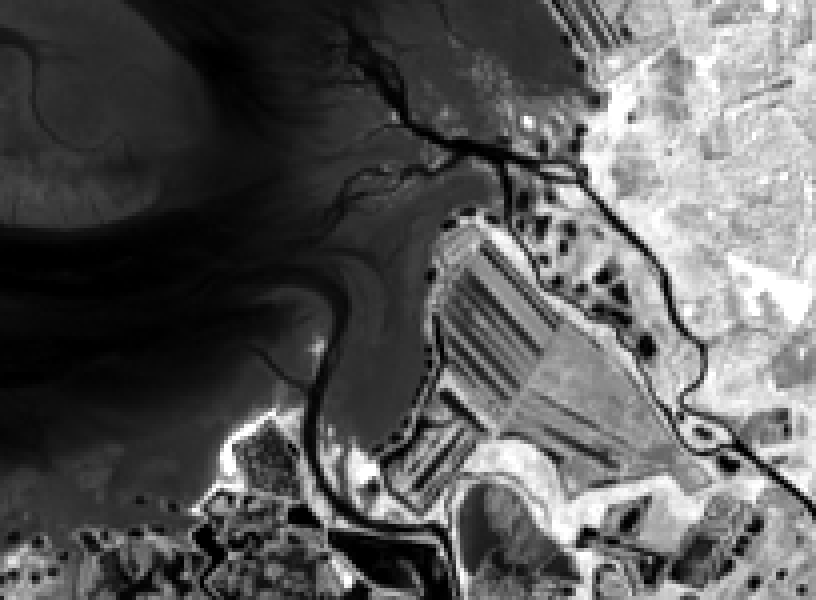}\includegraphics[width=0.5\textwidth]{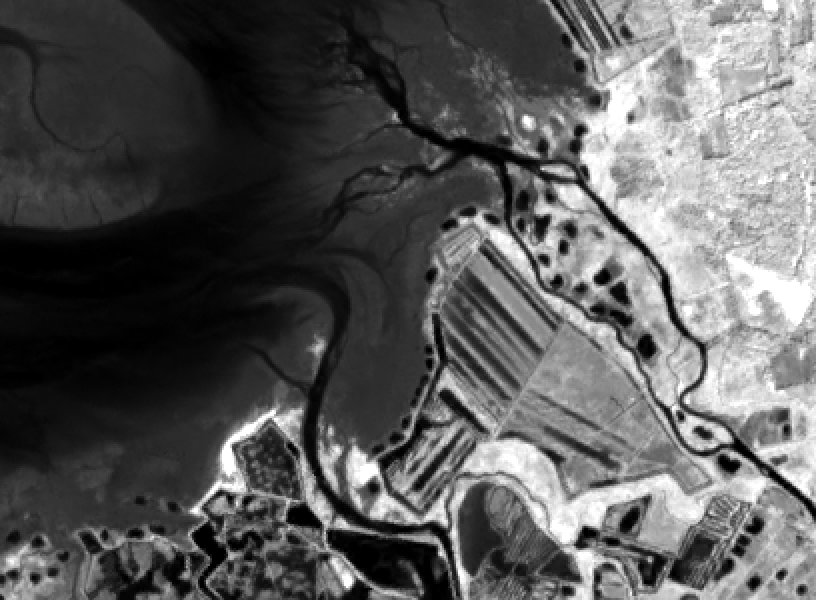}

\includegraphics[width=0.5\textwidth]{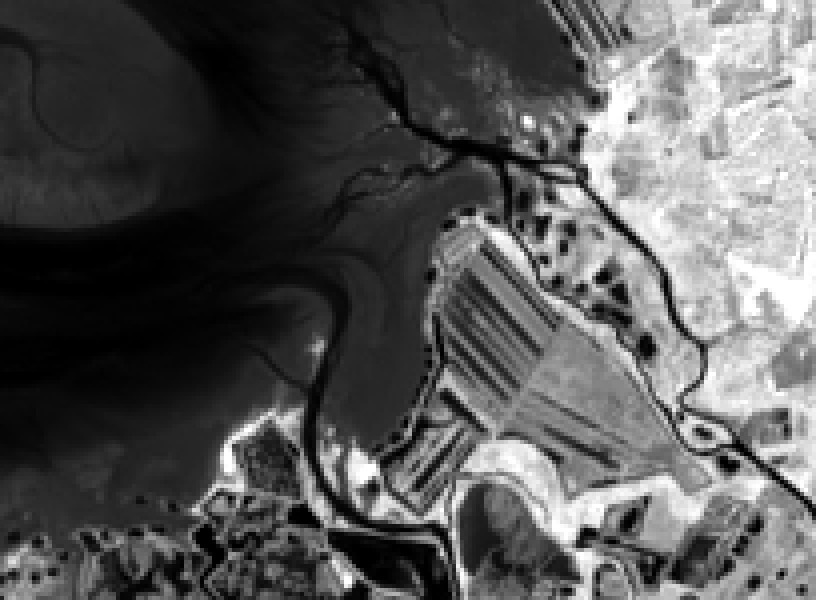}\includegraphics[width=0.5\textwidth]{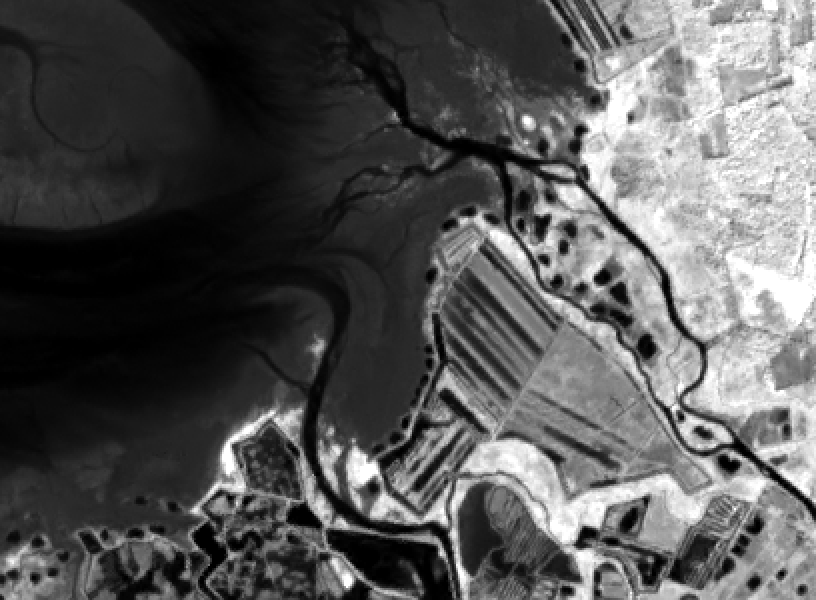}

\includegraphics[width=0.5\textwidth]{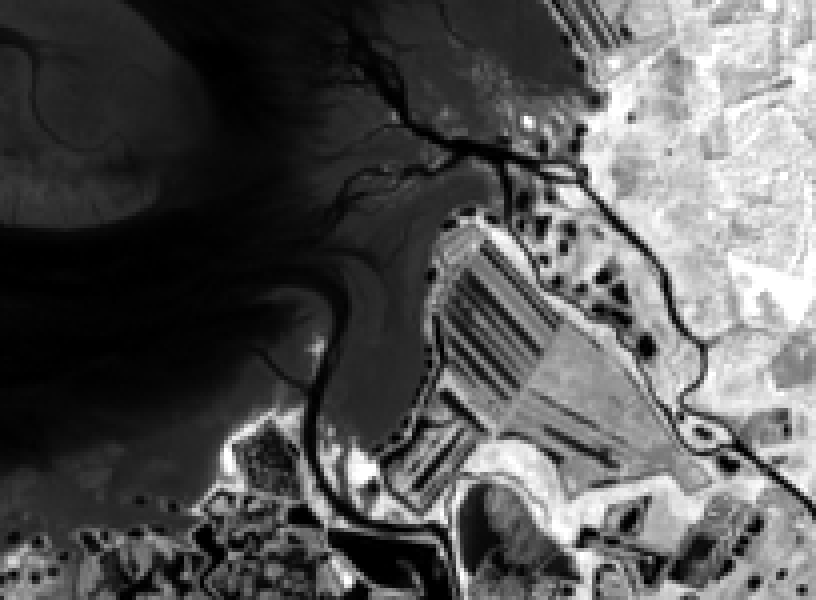}\includegraphics[width=0.5\textwidth]{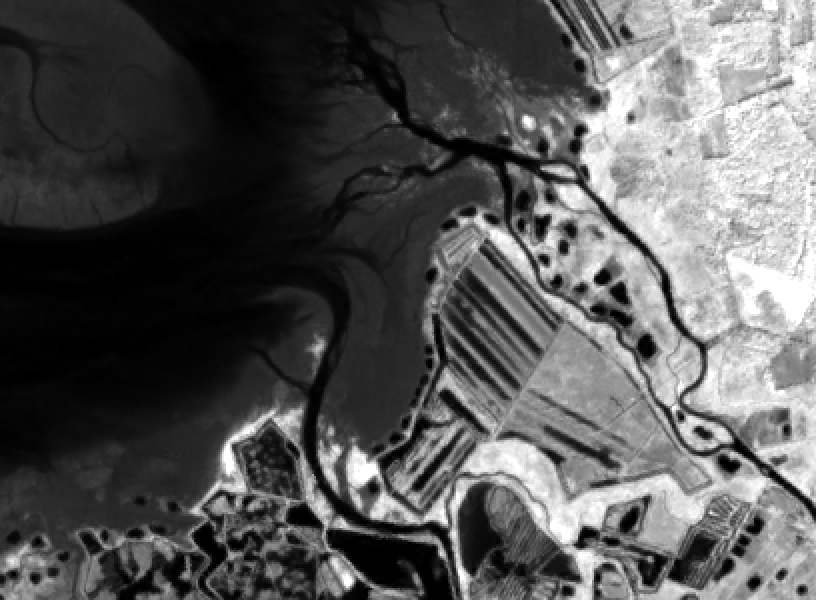}

\caption{\label{fig:Coastal_area_B6_B7_B8A}Coastal area (delta of the Leyre
river). Top-left: Original band 6 (close infrared, 740nm, 20m/pixel).
Top-right: Super-resolved band 6 at 10m/pixel. Middle-Left: Original
band 7 (close infrared, 783nm, 20m/pixel). Middle-Left: Super-resolved
band 7 at 10m/pixel. Bottom-left: Original band 8A (close infrared,
865nm, 20m/pixel). Bottom-right: Super-resolved band 8A at 10m/pixel.}
\end{figure*}

\begin{figure*}[b]
\includegraphics[width=0.5\textwidth]{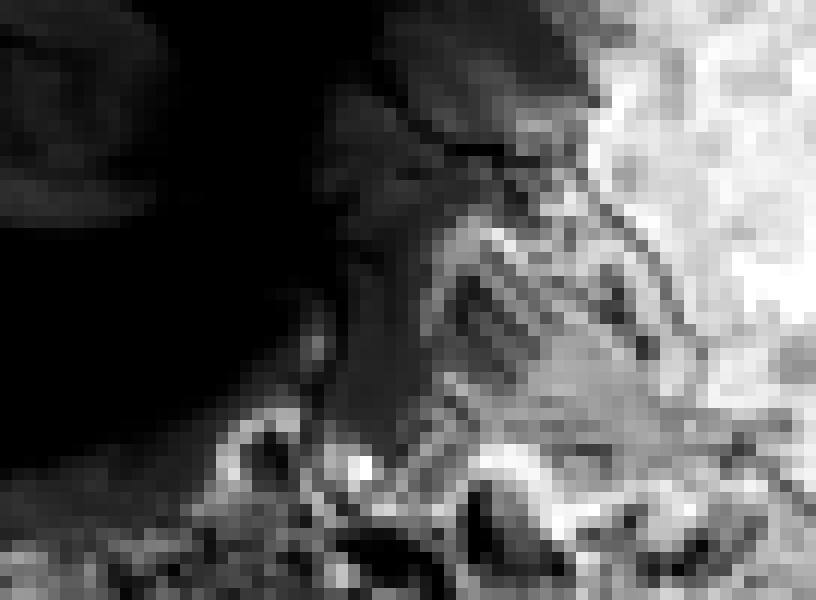}\includegraphics[width=0.5\textwidth]{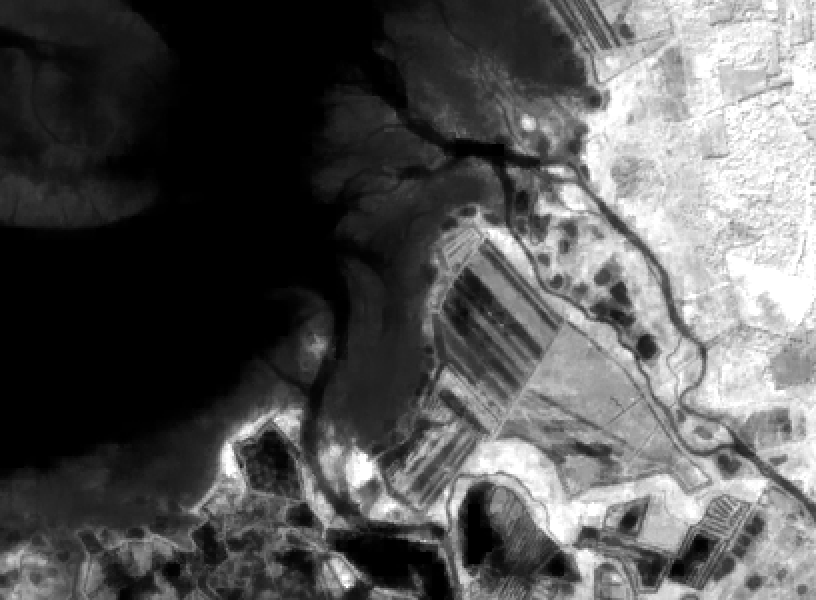}

\includegraphics[width=0.5\textwidth]{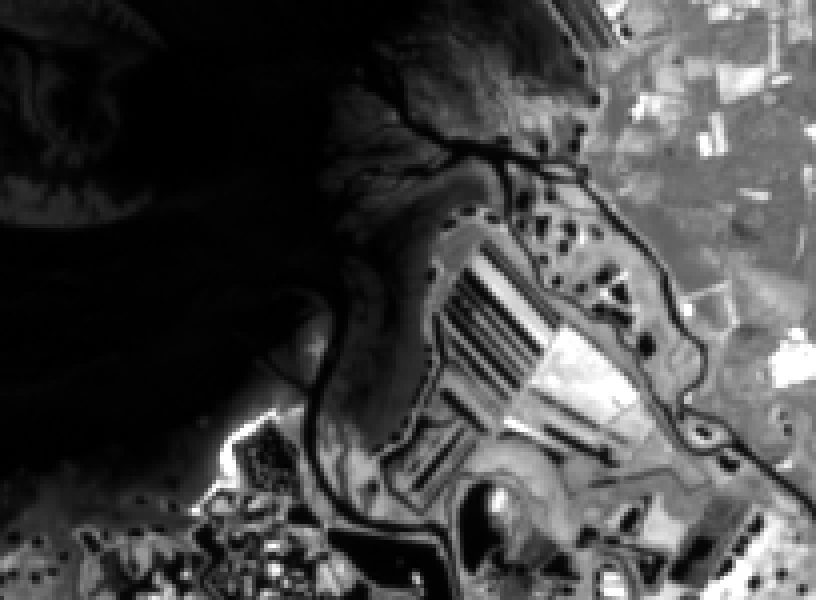}\includegraphics[width=0.5\textwidth]{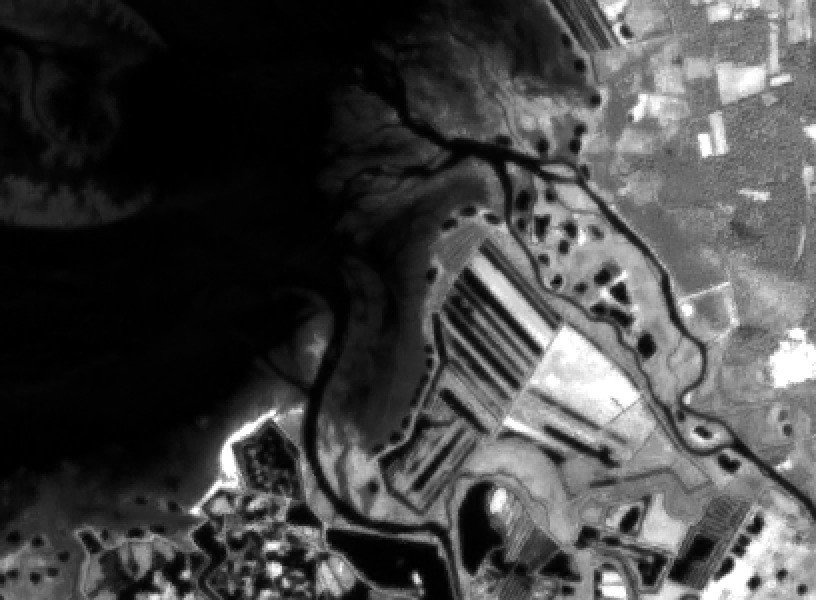}

\includegraphics[width=0.5\textwidth]{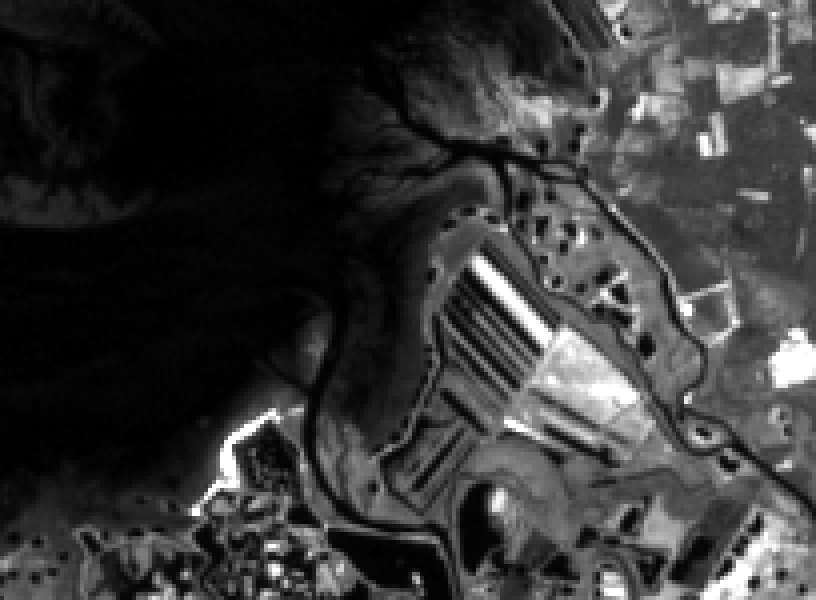}\includegraphics[width=0.5\textwidth]{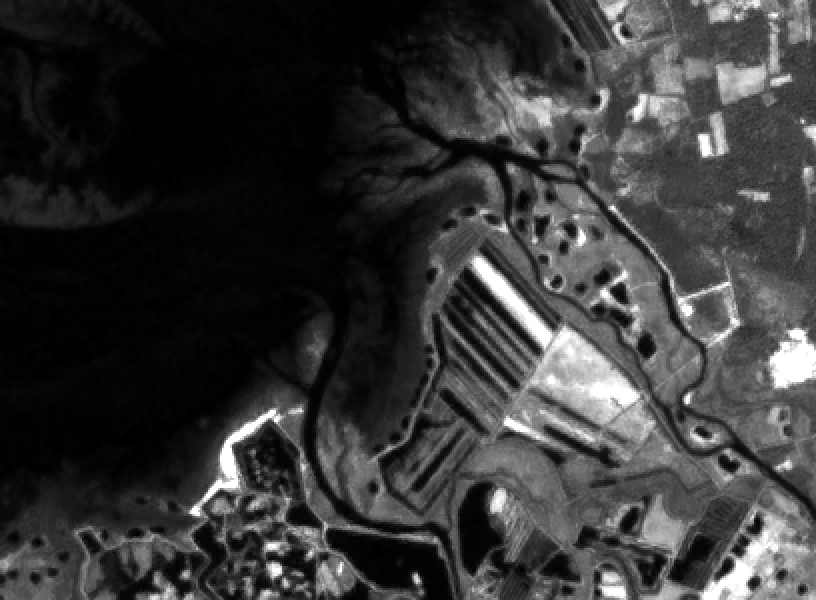}

\caption{\label{fig:Coastal_area_B9_B11_B12}Coastal area (delta of the Leyre
river). Top-left: Original band 9 (infrared, 945nm, 60m/pixel). Top-right:
Super-resolved band 9 at 10m/pixel. Middle-Left: Original band 11
(deep infrared, 1610nm, 20m/pixel). Middle-Left: Super-resolved band
11 at 10m/pixel. Bottom-left: Original band 12 (deep infrared, 2190nm,
20m/pixel). Bottom-right: Super-resolved band 12 at 10m/pixel.}
\end{figure*}

Results for the coastal environment are displayed in Figs.~\ref{fig:Coastal_area_RGB_B8_B1_B5},\ref{fig:Coastal_area_B6_B7_B8A},\ref{fig:Coastal_area_B9_B11_B12}.
Quantitative indicators for the 40m->20m super-resolution are :
\begin{center}
\begin{tabular}{|c||c|c|c|}
\hline 
Band & Q & ERGAS & SAM\tabularnewline
\hline 
\hline 
B5 (705nm) & 0.99 & 2.91 & 3.08\tabularnewline
\hline 
B6 (740nm) & 0.995 & 3.87 & 3.5\tabularnewline
\hline 
B7 (783nm) & 0.996 & 3.83 & 3.4\tabularnewline
\hline 
B11 (1610nm) & 0.994 & 5.19 & 4.26\tabularnewline
\hline 
B12 (2190nm) & 0.992 & 6.32 & 5.07\tabularnewline
\hline 
B8A (865nm) & 0.996 & 3.99 & 3.44\tabularnewline
\hline 
\hline 
Global & 0.994 & 4.49 & 3.79\tabularnewline
\hline 
\end{tabular}
\par\end{center}

\subsection{Urban Area}

\begin{figure*}[b]
\includegraphics[width=0.5\textwidth]{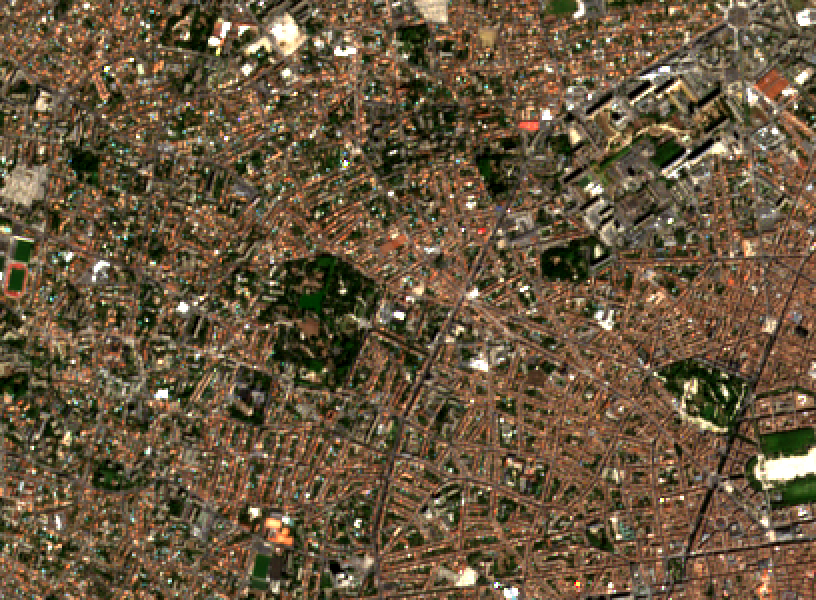}\includegraphics[width=0.5\textwidth]{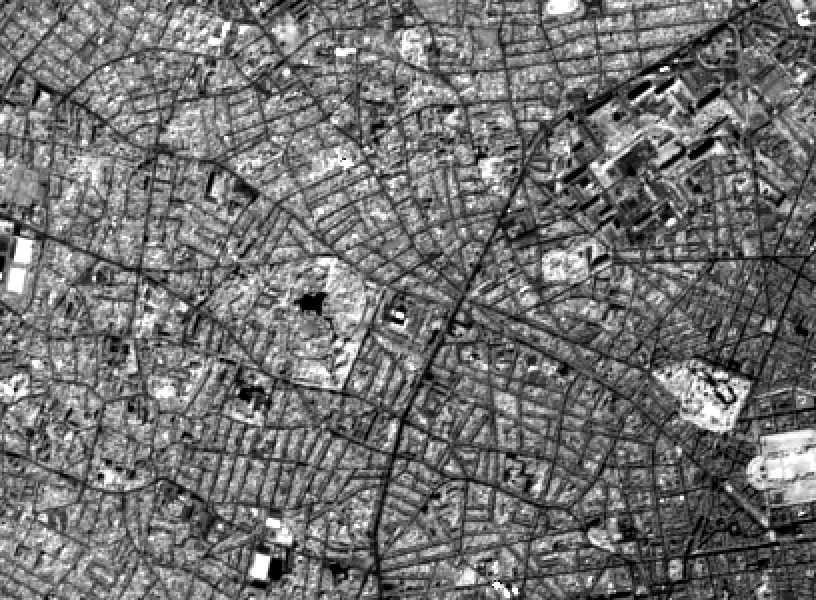}

\includegraphics[width=0.5\textwidth]{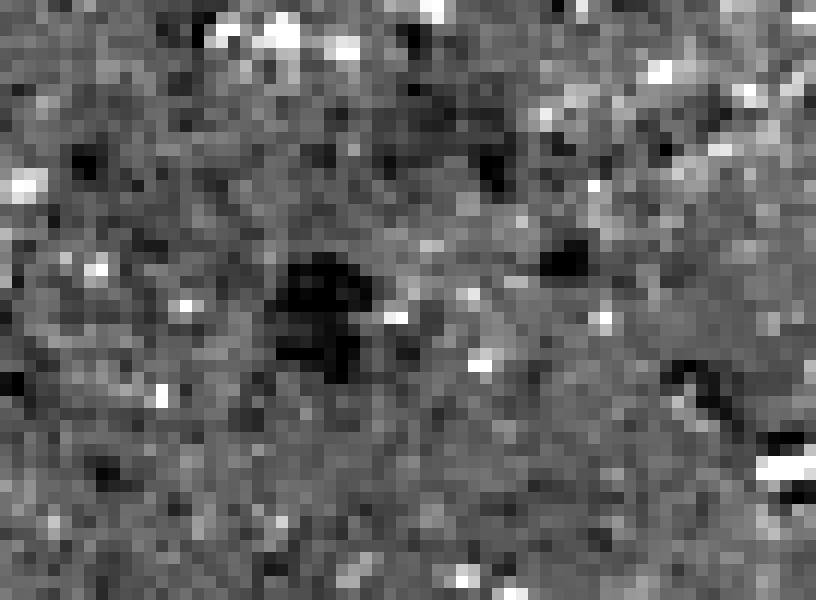}\includegraphics[width=0.5\textwidth]{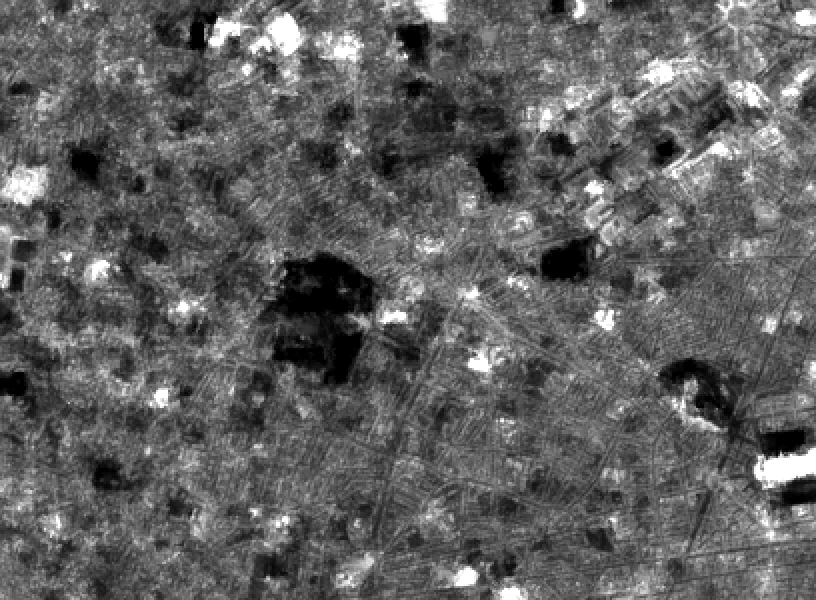}

\includegraphics[width=0.5\textwidth]{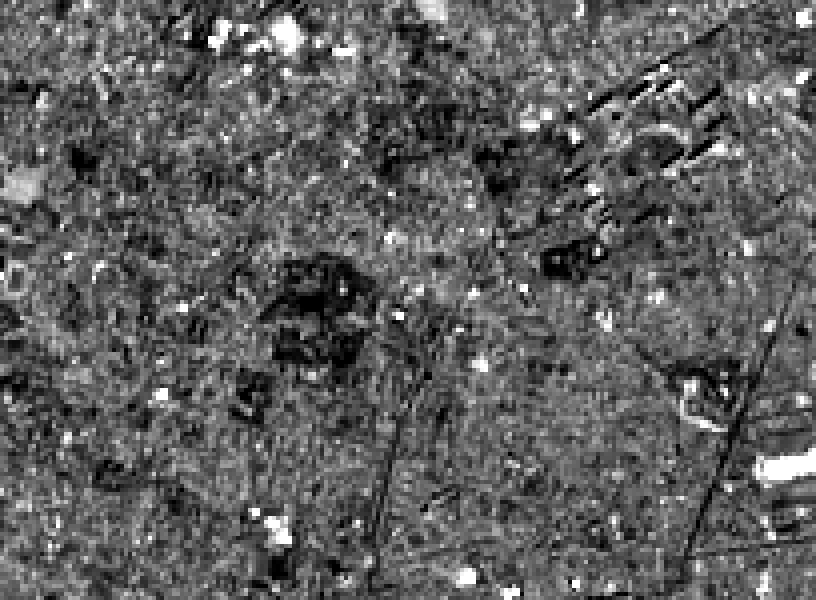}\includegraphics[width=0.5\textwidth]{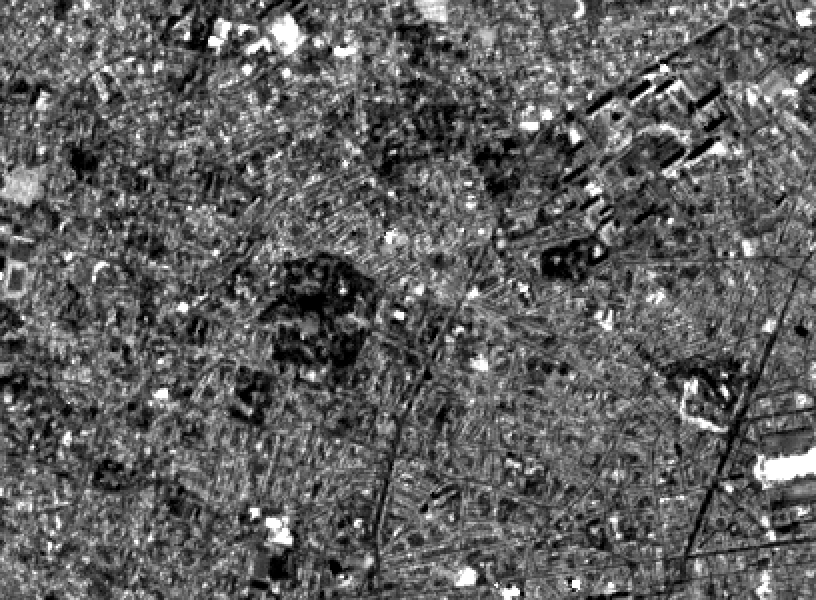}\caption{\label{fig:Urban_area_RGB_B8_B1_B5}Top-Left: Composite image of the
urban area (Bordeaux). Top-Right: Original infrared band 8 (842 nm,
10m/pixel). Middle-Left: Original band 1 (violet, 443nm, 60m/pixel).
Middle-Left: Super-resolved band 1 at 10m/pixel. Bottom-left: Original
band 5 (red-edge, 705nm, 20m/pixel). Bottom-right: Super-resolved
band 5 at 10m/pixel.}
\end{figure*}

\begin{figure*}[b]
\includegraphics[width=0.5\textwidth]{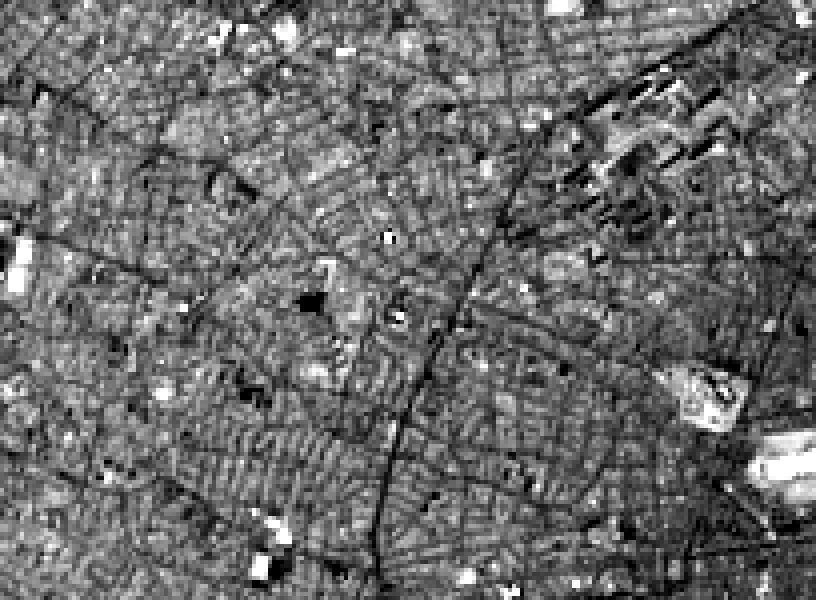}\includegraphics[width=0.5\textwidth]{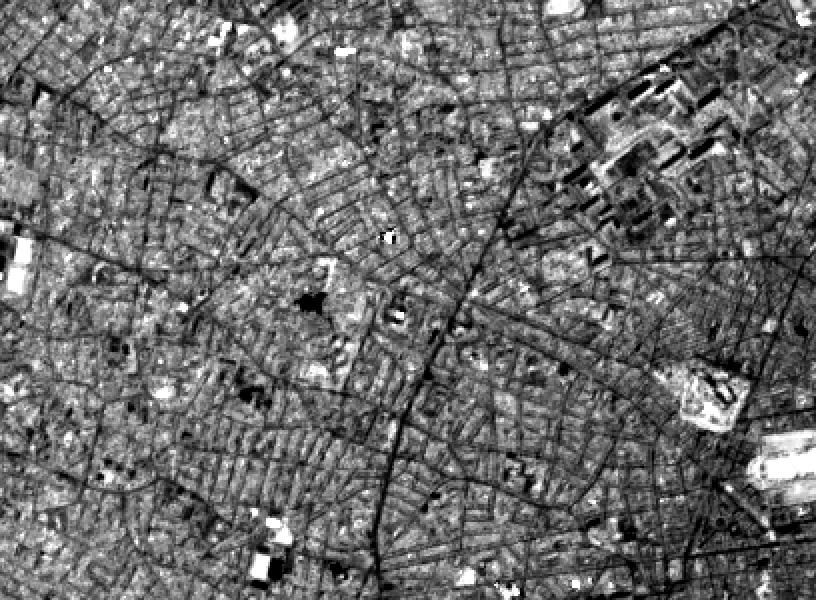}

\includegraphics[width=0.5\textwidth]{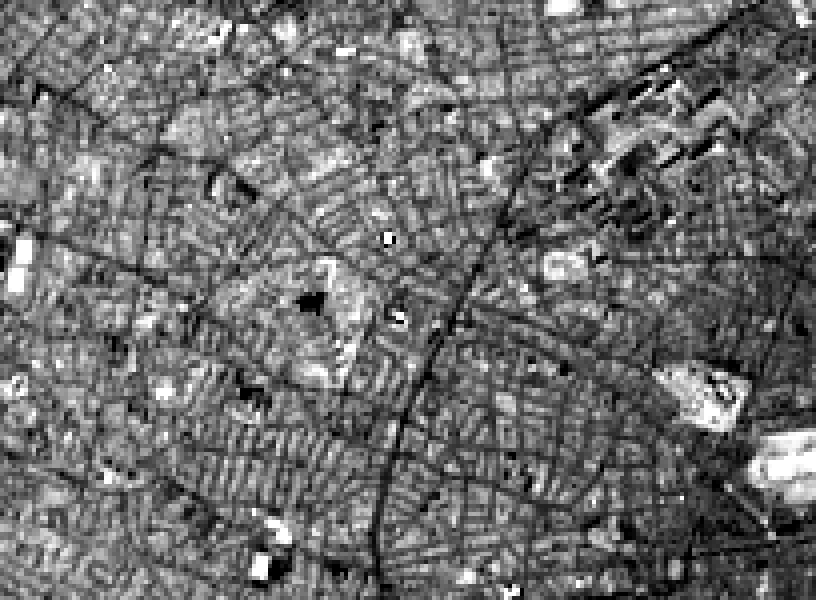}\includegraphics[width=0.5\textwidth]{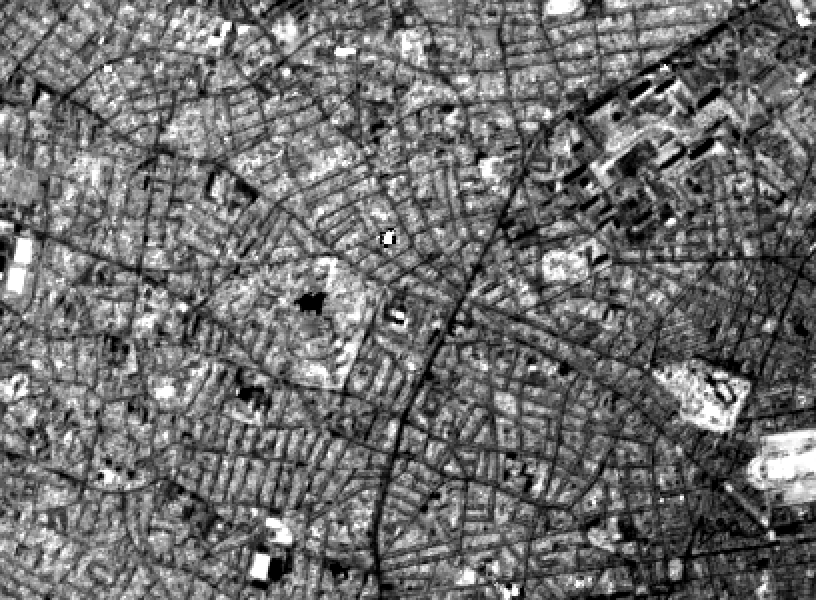}

\includegraphics[width=0.5\textwidth]{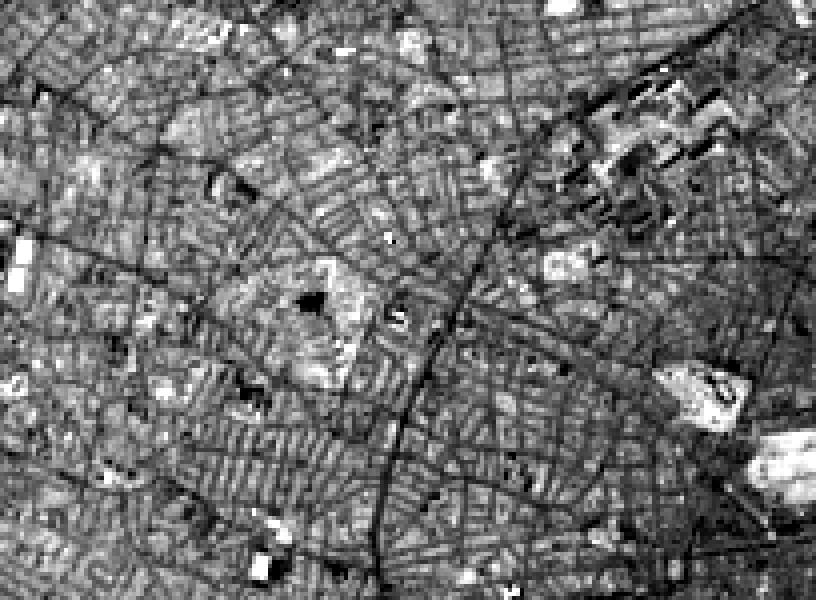}\includegraphics[width=0.5\textwidth]{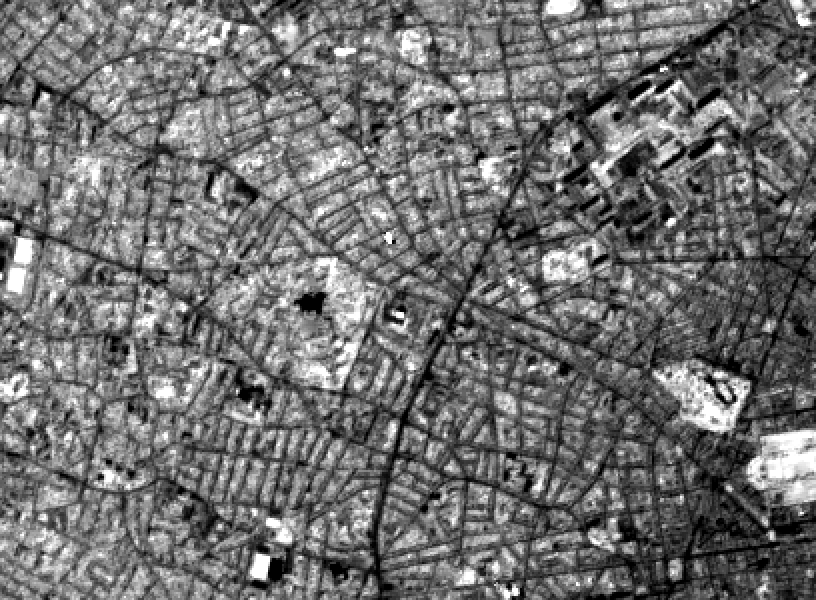}

\caption{\label{fig:Urban_area_B6_B7_B8A}Urban area (Bordeaux). Top-left:
Original band 6 (close infrared, 740nm, 20m/pixel). Top-right: Super-resolved
band 6 at 10m/pixel. Middle-Left: Original band 7 (close infrared,
783nm, 20m/pixel). Middle-Left: Super-resolved band 7 at 10m/pixel.
Bottom-left: Original band 8A (close infrared, 865nm, 20m/pixel).
Bottom-right: Super-resolved band 8A at 10m/pixel.}
\end{figure*}

\begin{figure*}[b]
\includegraphics[width=0.5\textwidth]{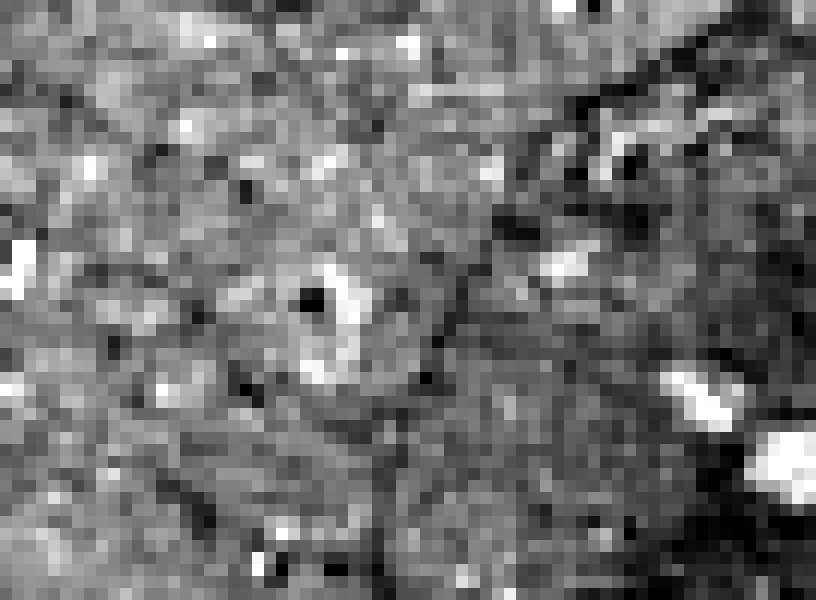}\includegraphics[width=0.5\textwidth]{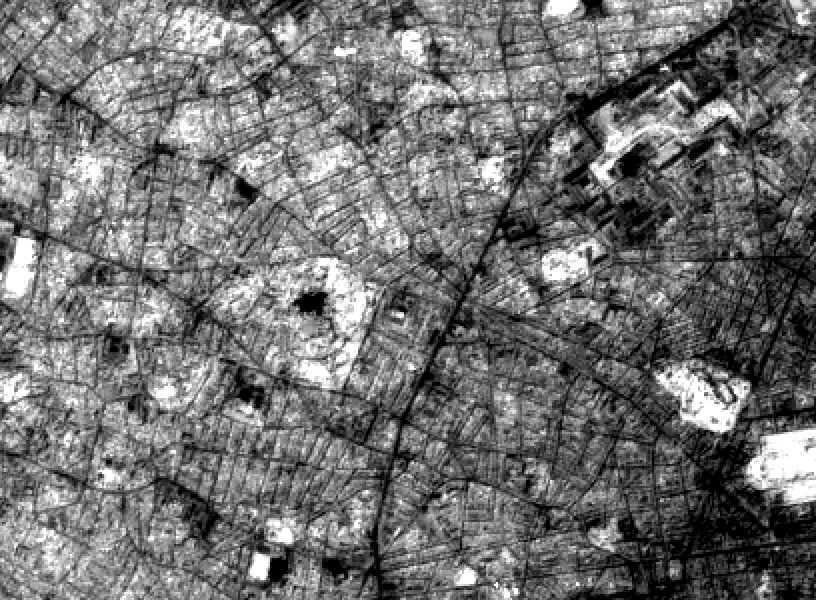}

\includegraphics[width=0.5\textwidth]{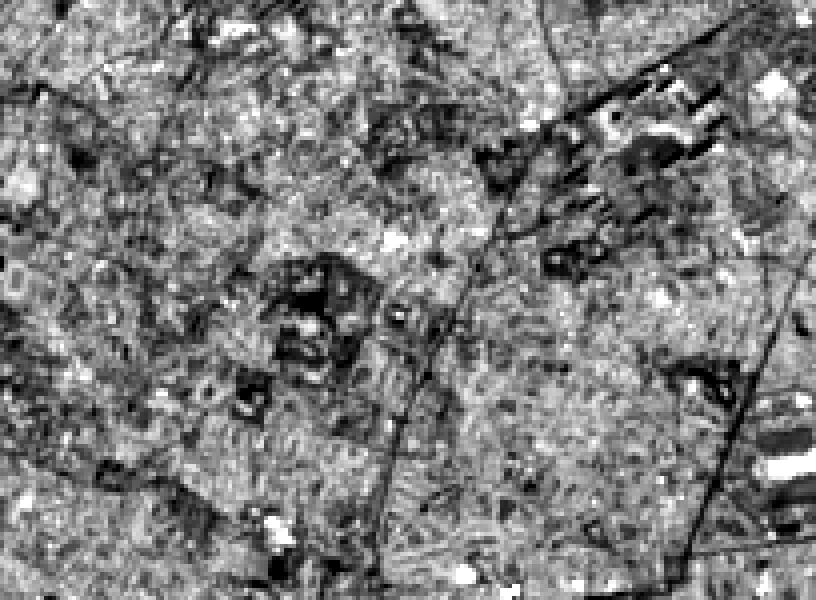}\includegraphics[width=0.5\textwidth]{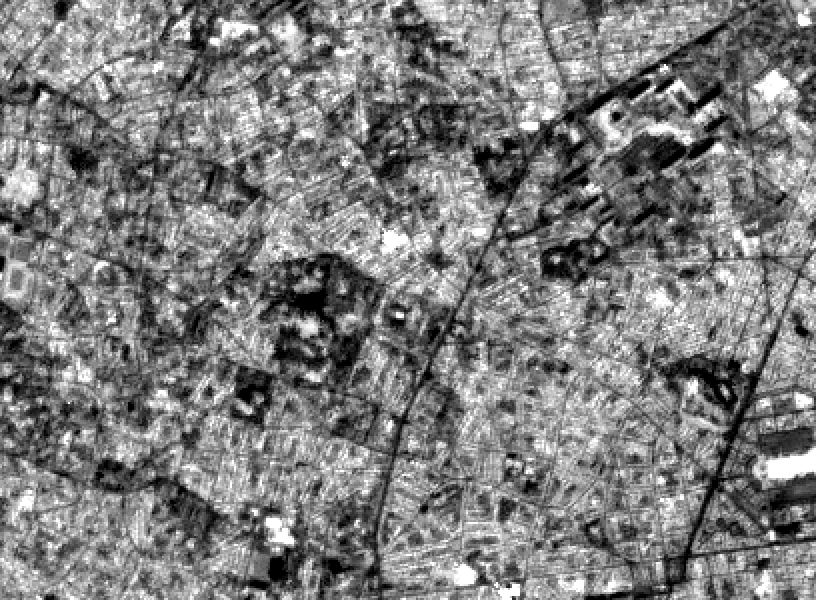}

\includegraphics[width=0.5\textwidth]{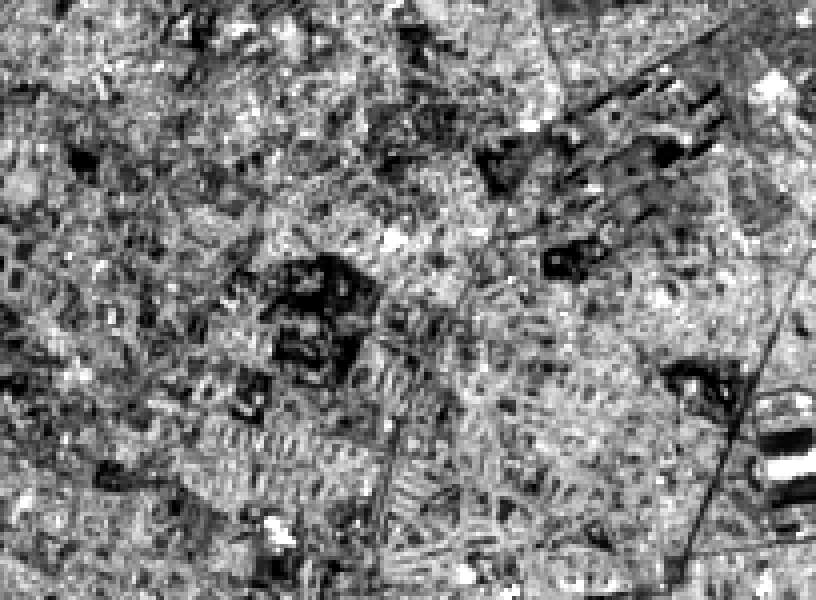}\includegraphics[width=0.5\textwidth]{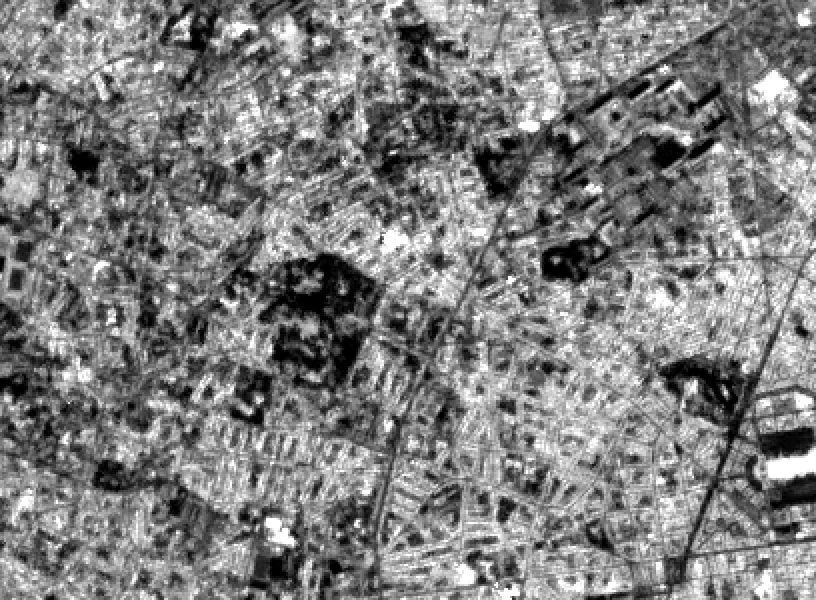}

\caption{\label{fig:Urban_area_B9_B11_B12}Urban area (Bordeaux). Top-left:
Original band 9 (infrared, 945nm, 60m/pixel). Top-right: Super-resolved
band 9 at 10m/pixel. Middle-Left: Original band 11 (deep infrared,
1610nm, 20m/pixel). Middle-Left: Super-resolved band 11 at 10m/pixel.
Bottom-left: Original band 12 (deep infrared, 2190nm, 20m/pixel).
Bottom-right: Super-resolved band 12 at 10m/pixel.}
\end{figure*}

Results for the urban environment are displayed in Figs.~\ref{fig:Urban_area_RGB_B8_B1_B5},\ref{fig:Urban_area_B6_B7_B8A},\ref{fig:Urban_area_B9_B11_B12}.
Quantitative indicators for the 40m->20m super-resolution are :
\begin{center}
\begin{tabular}{|c||c|c|c|}
\hline 
Band & Q & ERGAS & SAM\tabularnewline
\hline 
\hline 
B5 (705nm) & 0.942 & 4.98 & 5.47\tabularnewline
\hline 
B6 (740nm) & 0.948 & 3.94 & 4.38\tabularnewline
\hline 
B7 (783nm) & 0.95 & 4.07 & 4.51\tabularnewline
\hline 
B11 (1610nm) & 0.924 & 4.29 & 4.8\tabularnewline
\hline 
B12 (2190nm) & 0.928 & 5.34 & 5.89\tabularnewline
\hline 
B8A (865nm) & 0.956 & 3.76 & 4.17\tabularnewline
\hline 
\hline 
Global & 0.941 & 4.43 & 4.87\tabularnewline
\hline 
\end{tabular}
\par\end{center}

\subsection{Fields Area}

\begin{figure*}[b]
\includegraphics[width=0.5\textwidth]{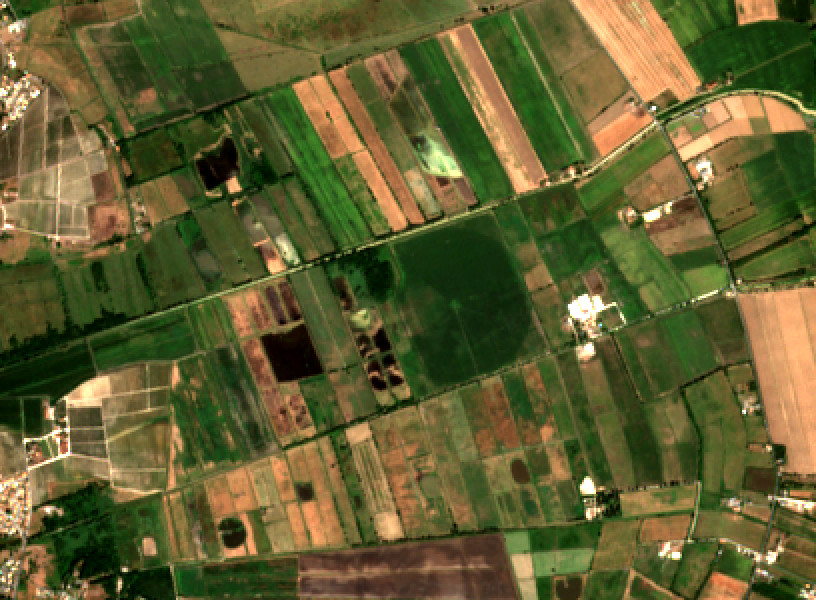}\includegraphics[width=0.5\textwidth]{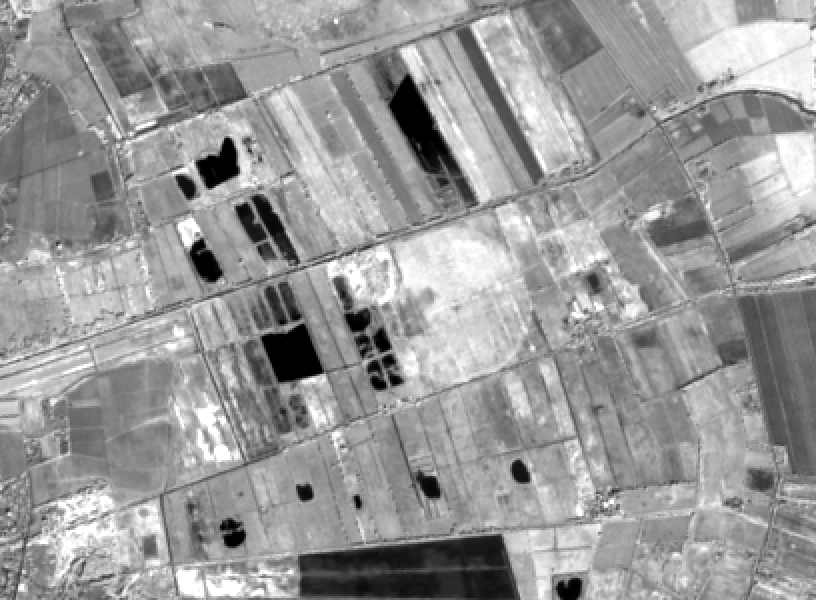}

\includegraphics[width=0.5\textwidth]{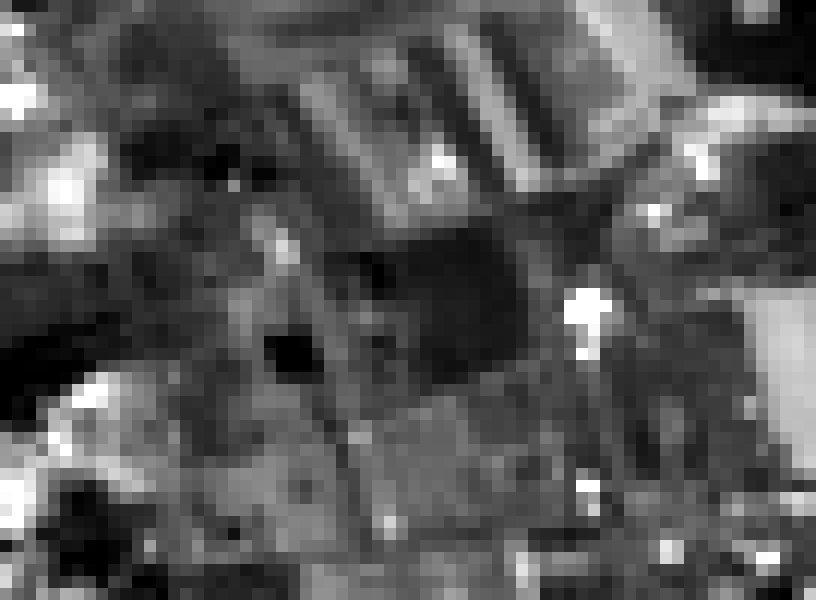}\includegraphics[width=0.5\textwidth]{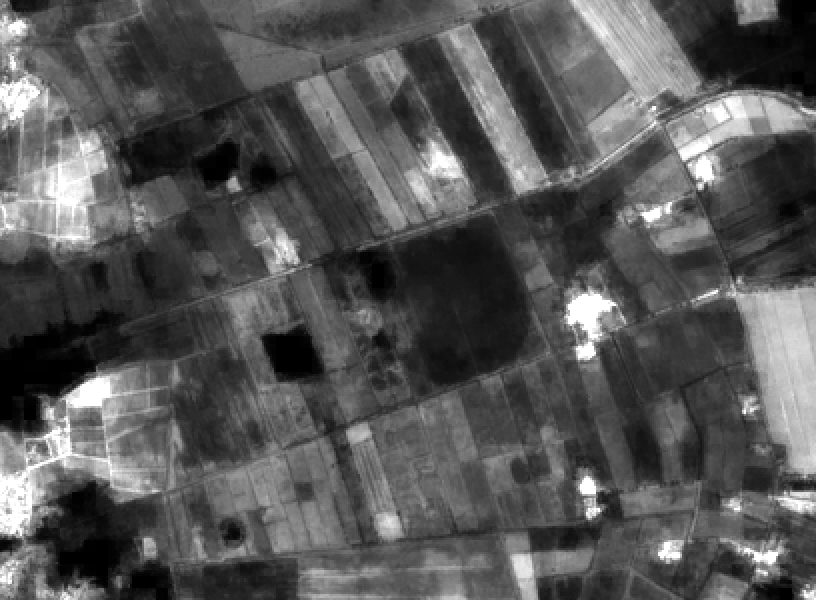}

\includegraphics[width=0.5\textwidth]{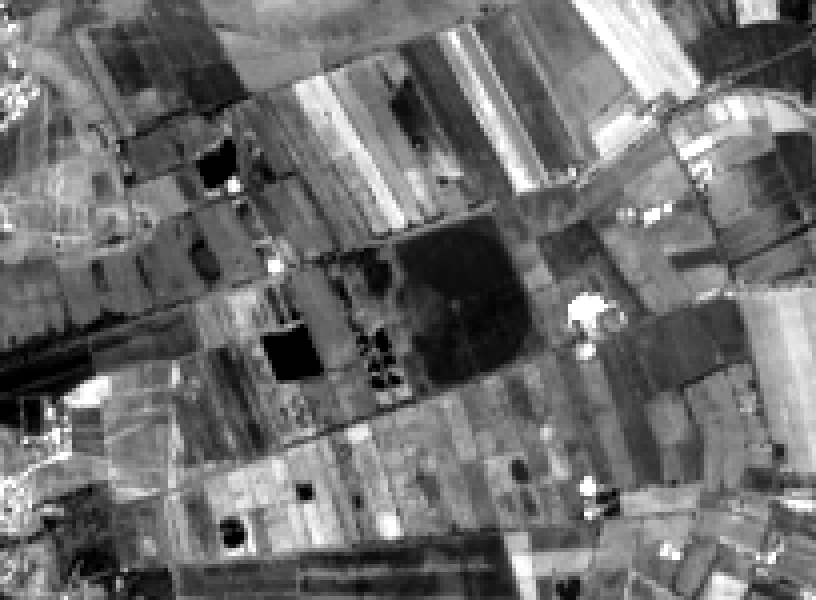}\includegraphics[width=0.5\textwidth]{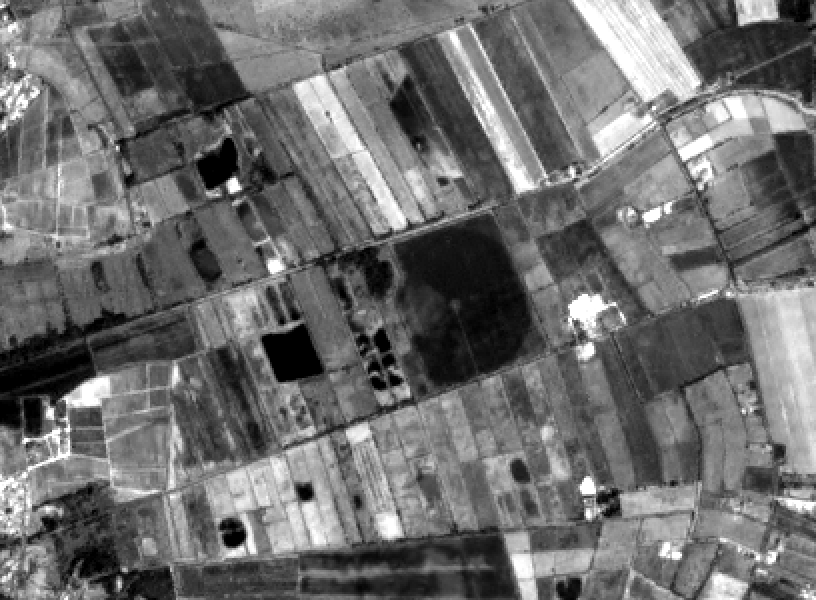}\caption{\label{fig:Fields_area_RGB_B8_B1_B5}Top-Left: Composite image of
the fields area (agriculture near bordeaux). Top-Right: Original infrared
band 8 (842 nm, 10m/pixel). Middle-Left: Original band 1 (violet,
443nm, 60m/pixel). Middle-Left: Super-resolved band 1 at 10m/pixel.
Bottom-left: Original band 5 (red-edge, 705nm, 20m/pixel). Bottom-right:
Super-resolved band 5 at 10m/pixel.}
\end{figure*}

\begin{figure*}[b]
\includegraphics[width=0.5\textwidth]{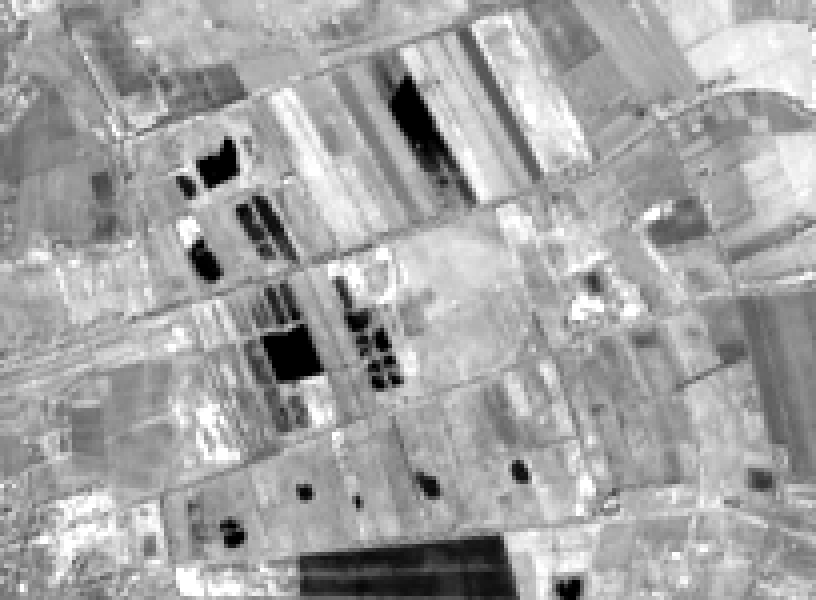}\includegraphics[width=0.5\textwidth]{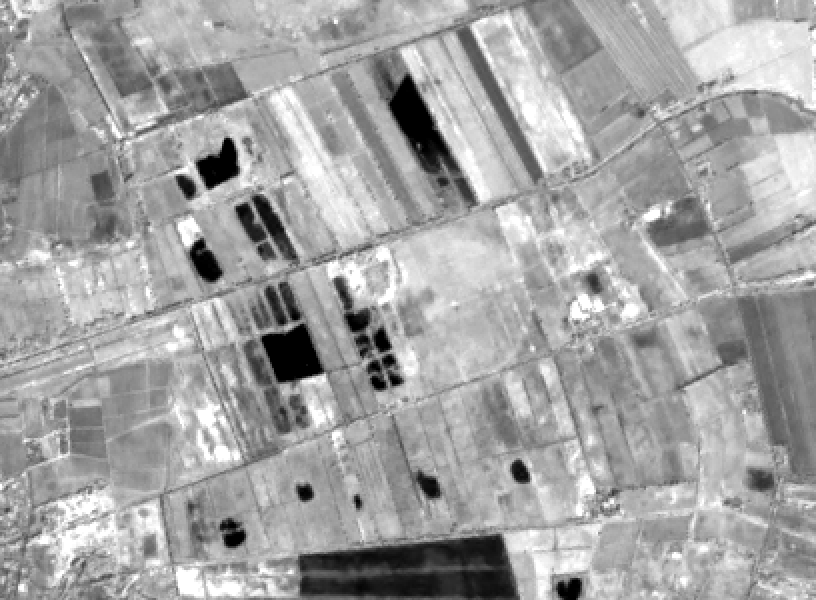}

\includegraphics[width=0.5\textwidth]{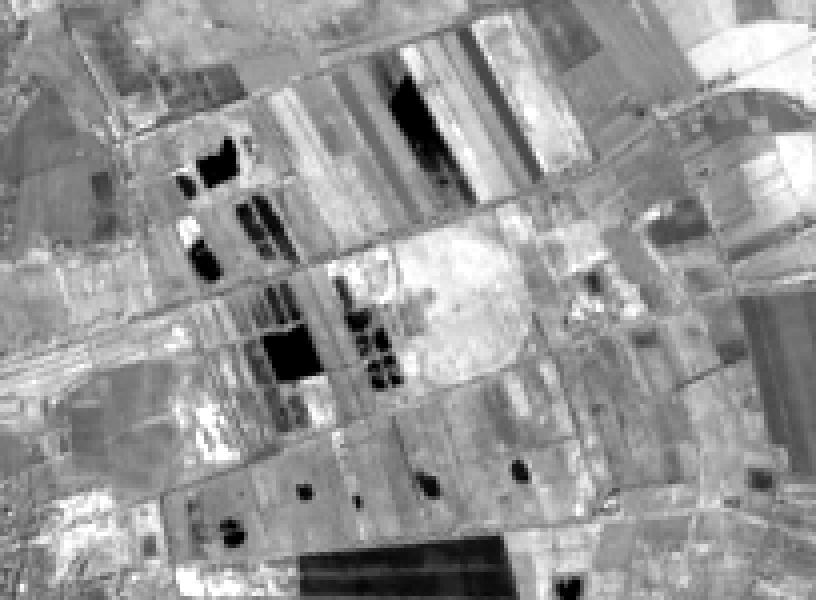}\includegraphics[width=0.5\textwidth]{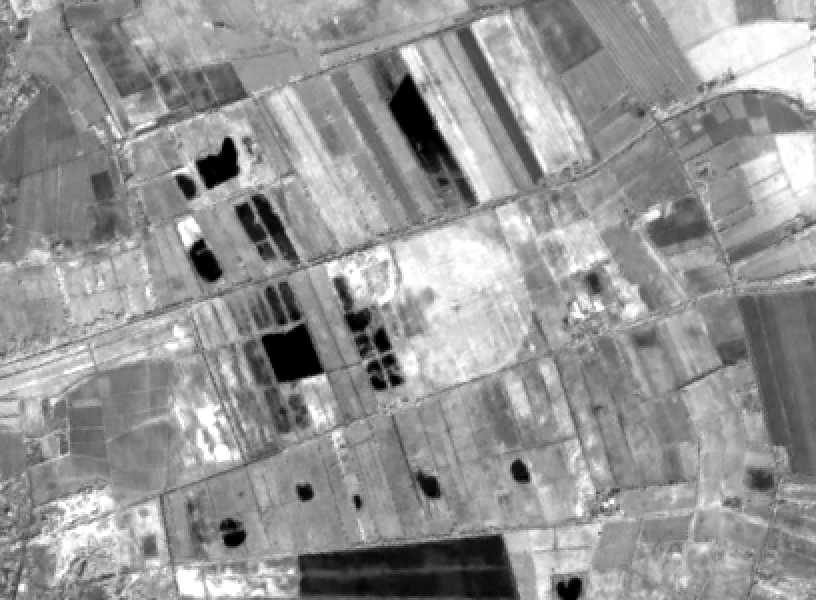}

\includegraphics[width=0.5\textwidth]{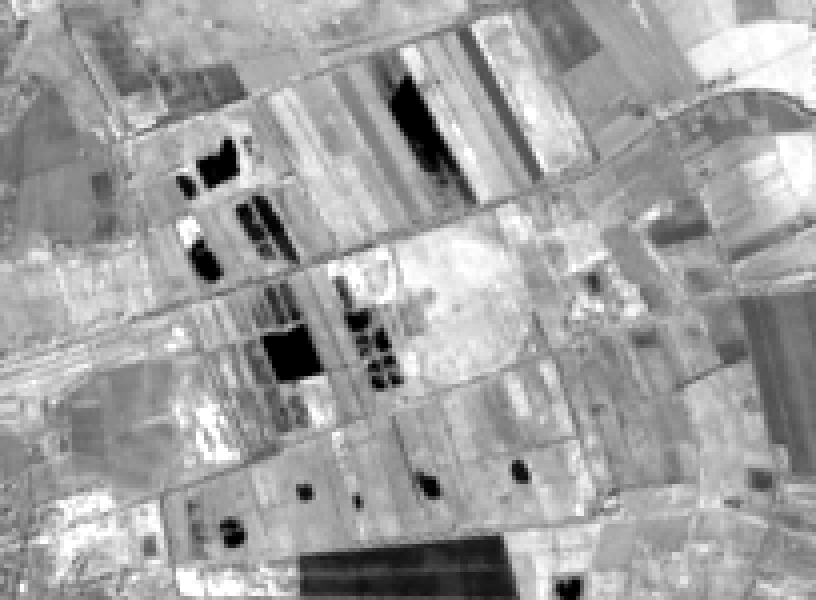}\includegraphics[width=0.5\textwidth]{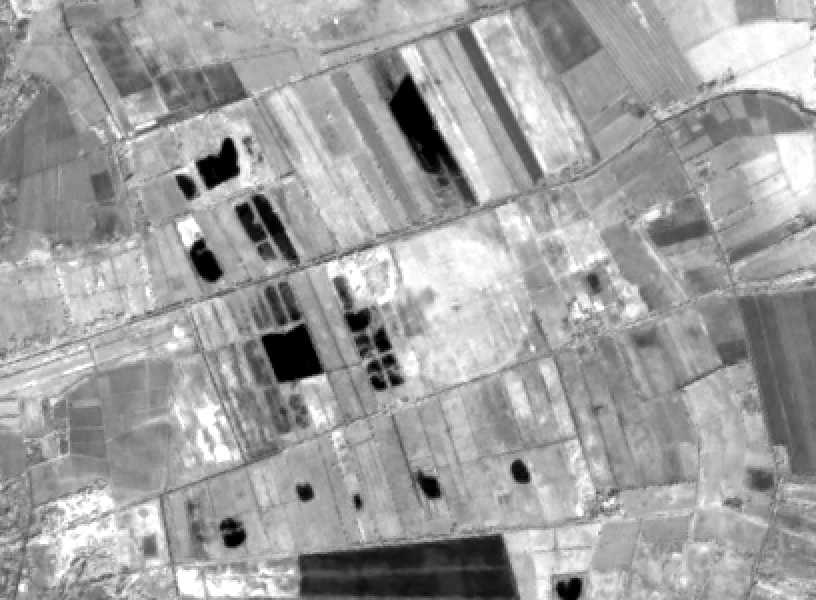}

\caption{\label{fig:Fields_area_B6_B7_B8A}Fields area (agriculture near Bordeaux).
Top-left: Original band 6 (close infrared, 740nm, 20m/pixel). Top-right:
Super-resolved band 6 at 10m/pixel. Middle-Left: Original band 7 (close
infrared, 783nm, 20m/pixel). Middle-Left: Super-resolved band 7 at
10m/pixel. Bottom-left: Original band 8A (close infrared, 865nm, 20m/pixel).
Bottom-right: Super-resolved band 8A at 10m/pixel.}
\end{figure*}

\begin{figure*}[b]
\includegraphics[width=0.5\textwidth]{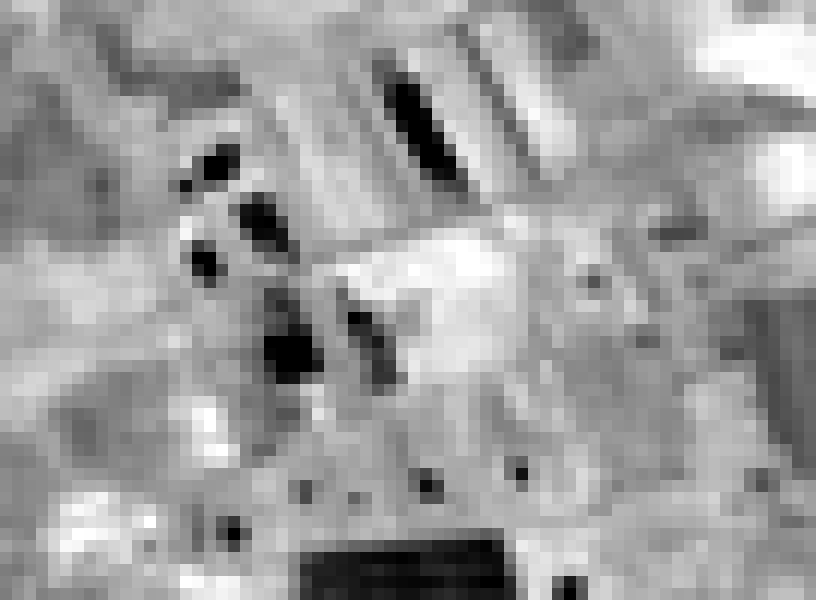}\includegraphics[width=0.5\textwidth]{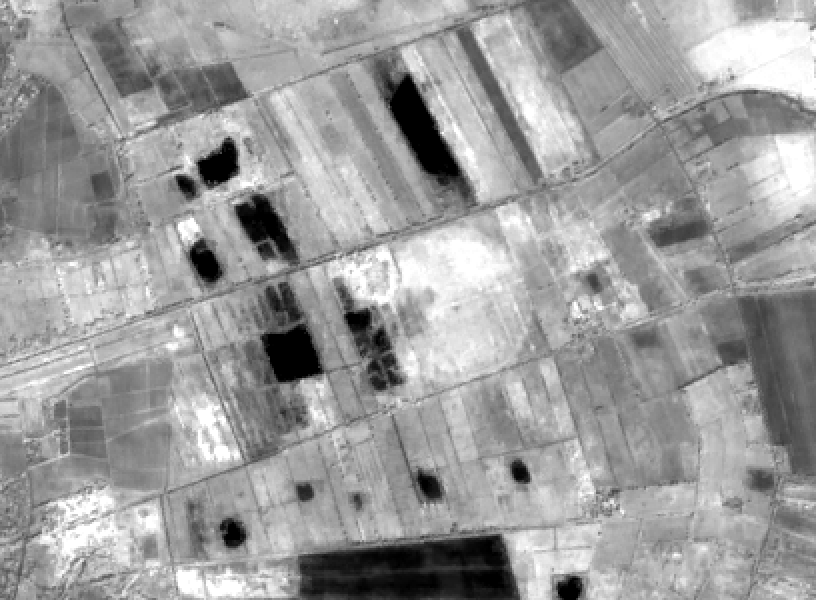}

\includegraphics[width=0.5\textwidth]{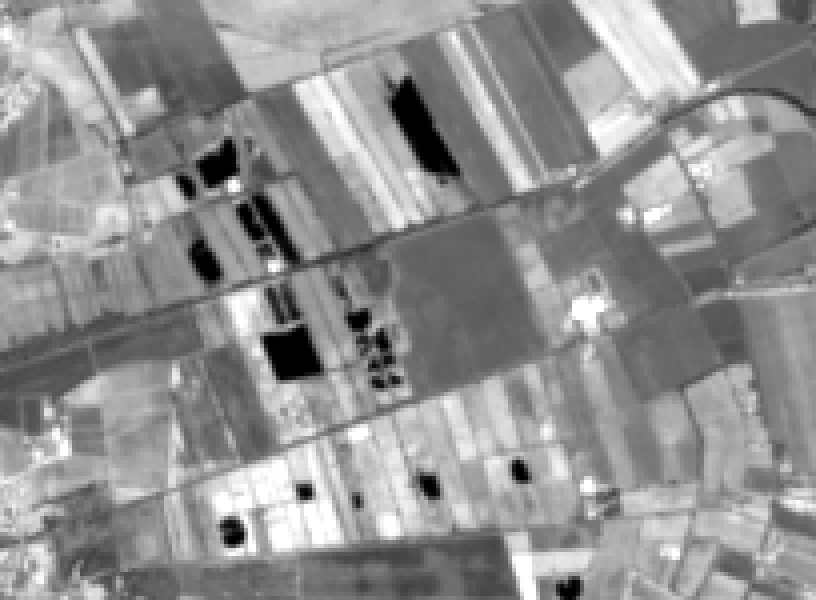}\includegraphics[width=0.5\textwidth]{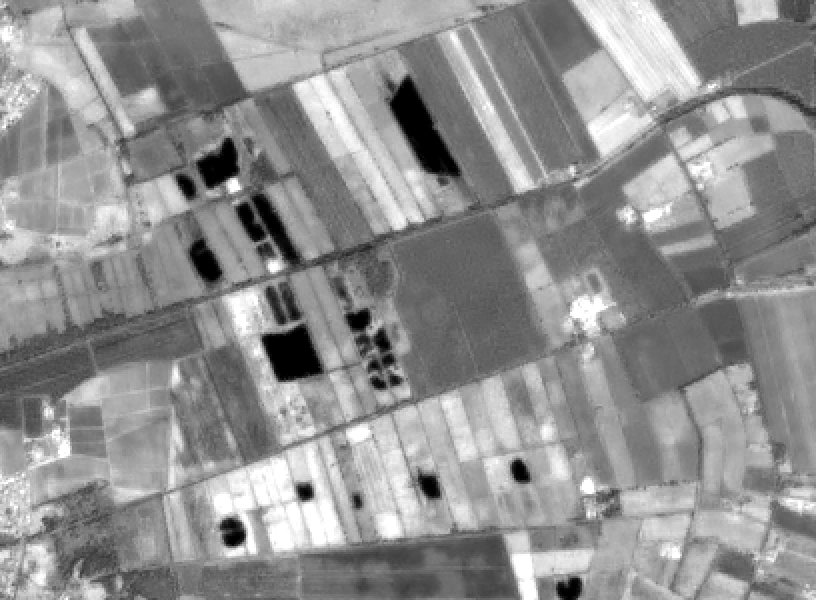}

\includegraphics[width=0.5\textwidth]{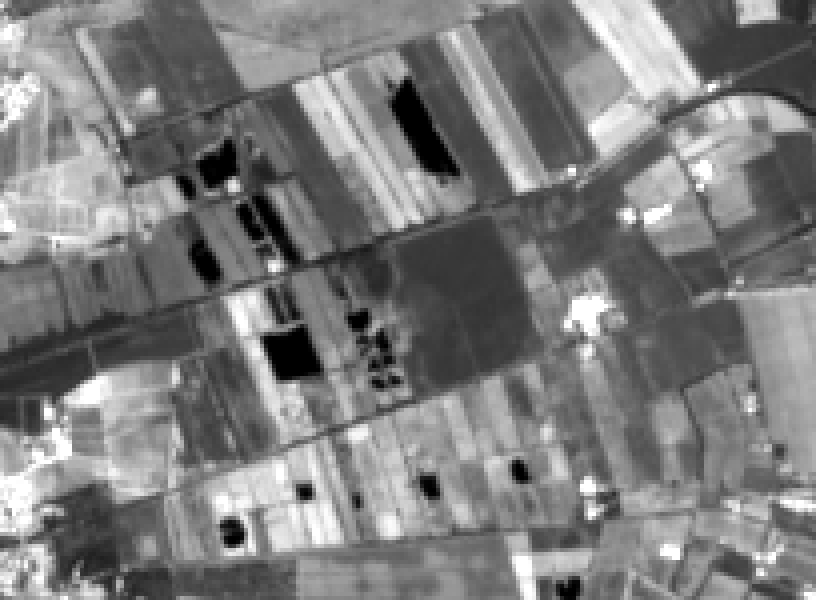}\includegraphics[width=0.5\textwidth]{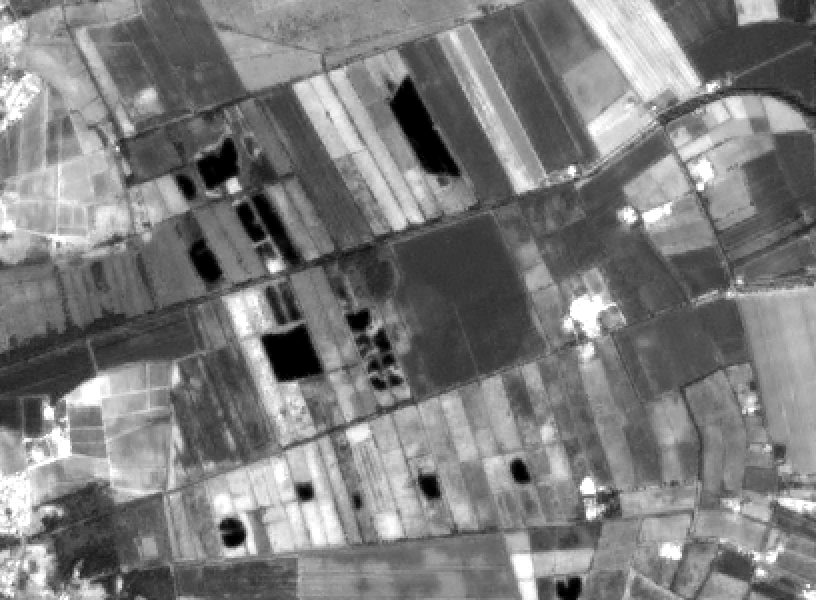}

\caption{\label{fig:Fields_area_B9_B11_B12}Fields area (agriculture near Bordeaux).
Top-left: Original band 9 (infrared, 945nm, 60m/pixel). Top-right:
Super-resolved band 9 at 10m/pixel. Middle-Left: Original band 11
(deep infrared, 1610nm, 20m/pixel). Middle-Left: Super-resolved band
11 at 10m/pixel. Bottom-left: Original band 12 (deep infrared, 2190nm,
20m/pixel). Bottom-right: Super-resolved band 12 at 10m/pixel.}
\end{figure*}

Results for the fields environment are displayed in Figs.~\ref{fig:Fields_area_RGB_B8_B1_B5},\ref{fig:Fields_area_B6_B7_B8A},\ref{fig:Fields_area_B9_B11_B12}.
Quantitative indicators for the 40m->20m super-resolution are :
\begin{center}
\begin{tabular}{|c||c|c|c|}
\hline 
Band & Q & ERGAS & SAM\tabularnewline
\hline 
\hline 
B5 (705nm) & 0.99 & 2.34 & 2.54\tabularnewline
\hline 
B6 (740nm) & 0.991 & 1.61 & 1.79\tabularnewline
\hline 
B7 (783nm) & 0.994 & 1.57 & 1.74\tabularnewline
\hline 
B11 (1610nm) & 0.988 & 2.3 & 2.53\tabularnewline
\hline 
B12 (2190nm) & 0.989 & 2.81 & 3.01\tabularnewline
\hline 
B8A (865nm) & 0.994 & 1.47 & 1.63\tabularnewline
\hline 
\hline 
Global & 0.991 & 2.08 & 2.21\tabularnewline
\hline 
\end{tabular}
\par\end{center}

\section{\label{sec:Discussion}Discussion}

Details not present in the original bands are immediately visible
in all images, especially at the largest super-resolution 60m->10m
(e.g. Fig \ref{fig:Coastal_area_RGB_B8_B1_B5}, middle-right). These
details correspond to the band-independant information that was extracted
from the other bands, and propagated to these images. Although each
band presents different reflectance properties (in particular, B1
and B9), the exact same weights and geometric information extracted
from Eqs.~\ref{eq:solve_for_weights},\ref{eq:fitting_V_from_LR},\ref{eq:average_ratio}
were applied to all super-resolved images. This example demonstrates
how the method correctly extracts band-independant information that
encodes image details, while preserving the reflectance of each band
(Eqs.~\ref{eq:L_eq_Havg},\ref{eq:L_eq_Havg_x9}).

Quantitative indicators are given for three typical land cover types.
The method works best for agricultural environments, with large uniform
areas, and worst in urban environments. Even then, Fig.~\ref{fig:Urban_area_B9_B11_B12},
right, shows details that are very well recovered. Comparison with
quantitative indicators from other works using panchromatic sharpening
should be taken with caution. We estimate only the 40m->20m super-resolution
proxy for reasons mentionned in Section \ref{sec:Performance-assessement}.
Nevertheless, compared to the litterature \cite{Wald00,Alparone08,Wu15},
our global averages of $0.941\leq Q\leq0.994$, $2.08\leq ERGAS\leq4.49$
and $2.21\leq SAM\leq4.87$ in the different regions of interest for
Sentinel-2 data are consistent with the state of art for panchromatic
sharpening, albeit without a panchromatic band.

\begin{figure*}
\includegraphics[width=0.5\textwidth]{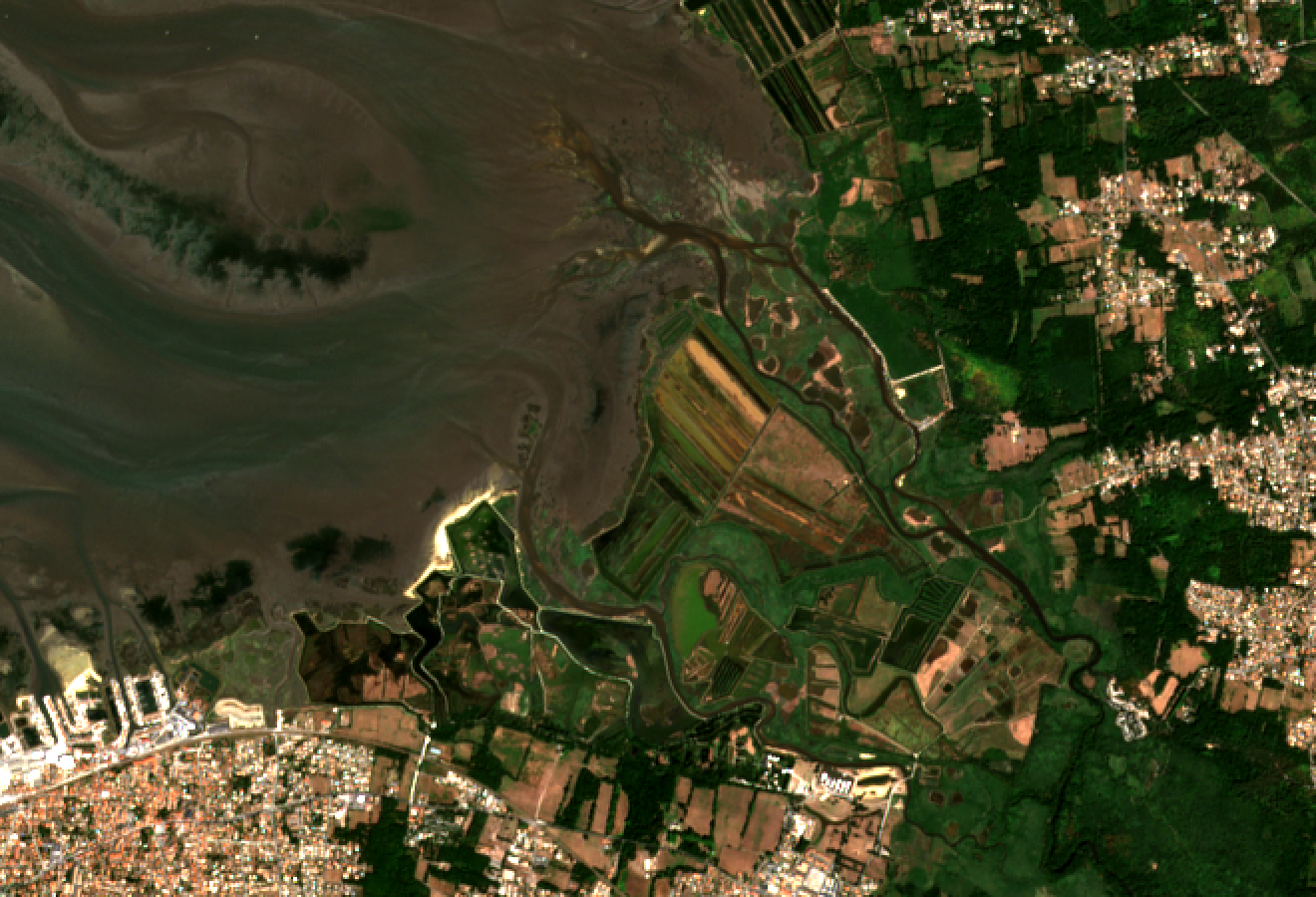}\includegraphics[width=0.5\textwidth]{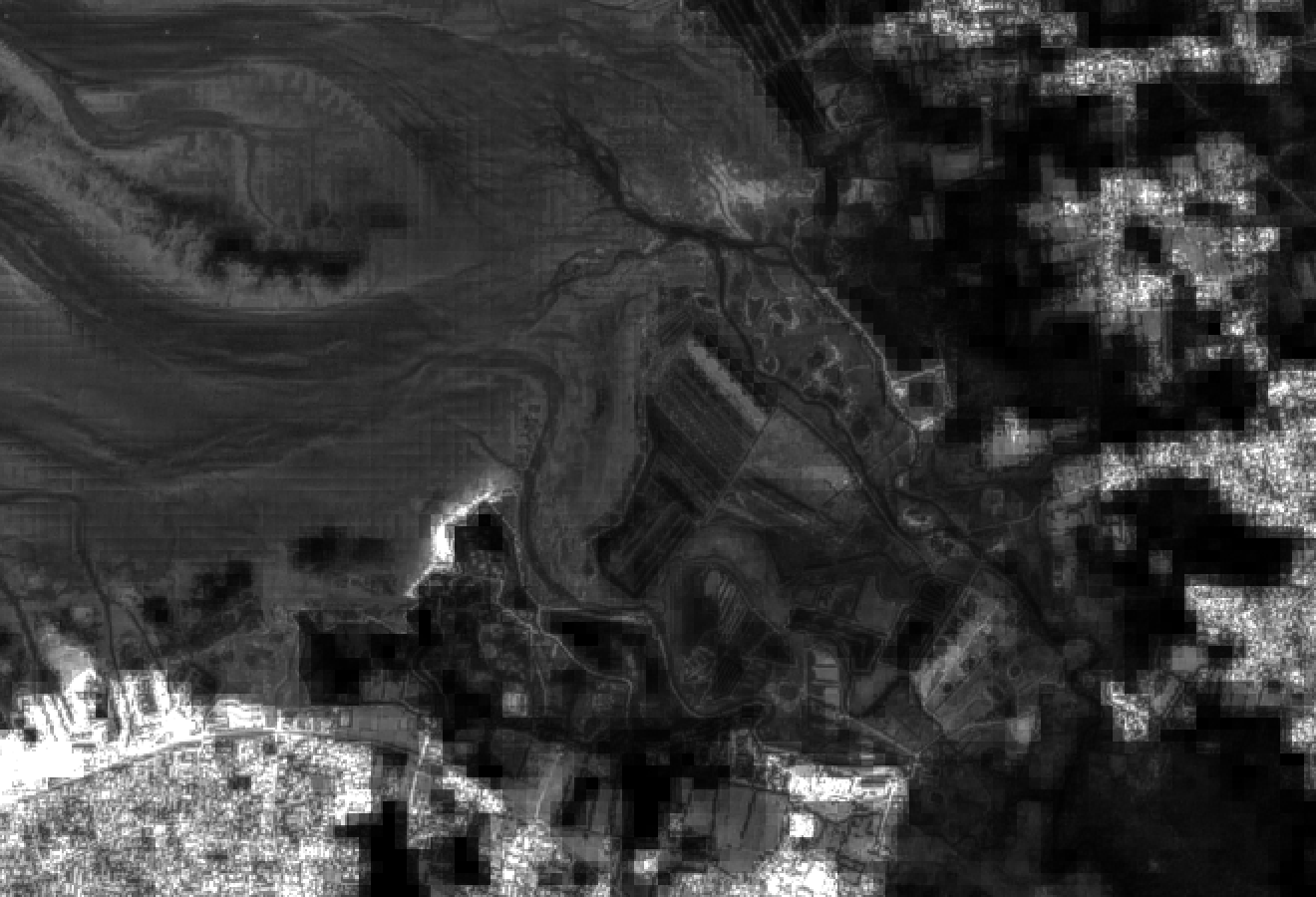}

\includegraphics[width=0.5\textwidth]{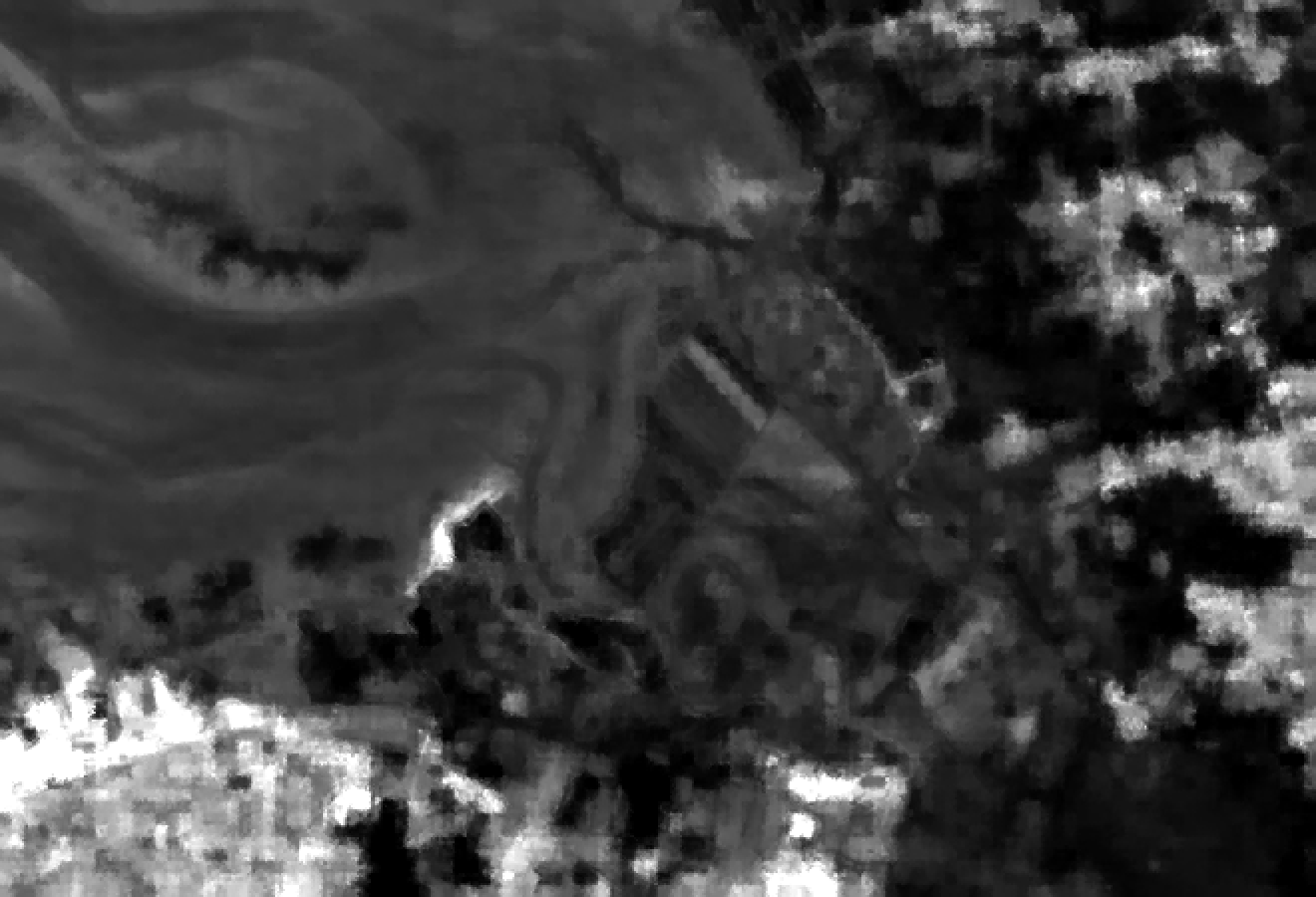}\includegraphics[width=0.5\textwidth]{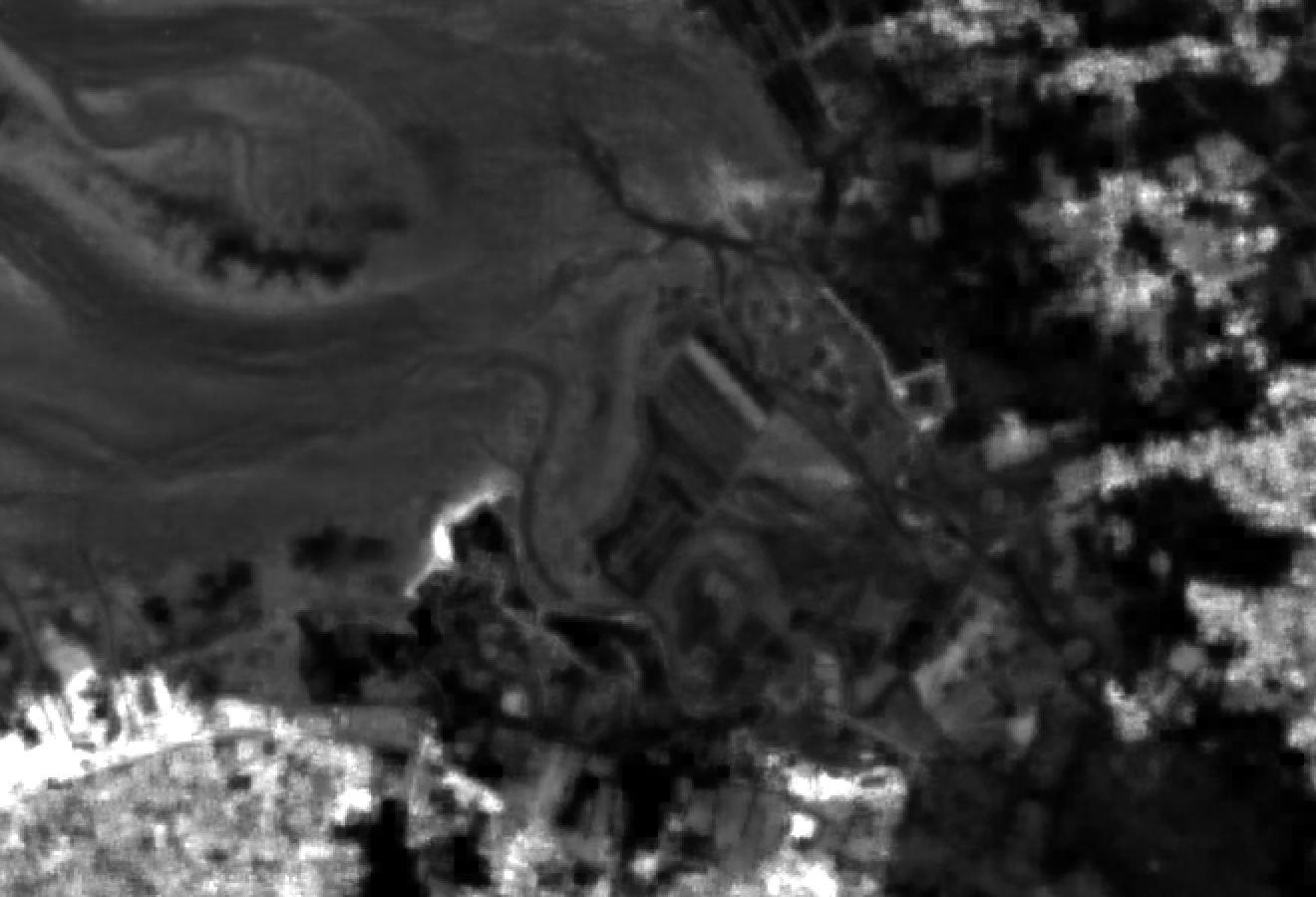}

\caption{\label{fig:Leyre_tests}Top-left: RGB composite image of the enlarged
coastal area, including fields and cities. Top-right: Incorrect image
obtained when shared values are not used, that demonstrates their
role in maintaining spatial consistency. Bottom-left: Incorrect image
obtained when only mean square optimisation is applied, that shows
the importance of the $\bar{q}$ ratio sharpening step described in
Section~\ref{subsec:Estimating-shared-values}. Bottom-right: Incorrect
image obtained when the estimation of weights $W$ is omitted while
maintaning the $\bar{q}$ ratio sharpening step. }
\end{figure*}

The method presented here includes multiple steps that are not trivial,
but they are all necessary. In order to demonstrate this, a final
experiment is performed on the coastal area, enlarged to encompass
nearby cities and fields (Fig.~\ref{fig:Leyre_tests}). The 60m band
1 is used in order to best visualize the effect of each step. A fist
idea would be to apply ratio sharpening (Eq.~\ref{eq:average_ratio})
directly on data values, instead of spatially shared values, so as
to simulate panchromatic sharpening. This would simplify the method
drastically. The result of this experiment is shown in Fig.~\ref{fig:Leyre_tests},
top-right, to be compared with the correct super-resolved result in
Fig.~\ref{fig:Coastal_area_RGB_B8_B1_B5}, middle-right. The role
of spatial consistenty is immediately highlighted: without the shared
values trick, unacceptable pixel blocks are clearly visible in the
result image. Conversely, why is $\bar{q}$ ratio sharpening useful?
Fig.~\ref{fig:Leyre_tests}, bottom-left shows that it is in fact
quite important for recovering the fine structures. Given that importance,
one may then question the usefulness of extracting weights $W$ as
band-independant information, especially since we also compute reverse
weights $V$ in a second step. Why would these $W$ encode image details?
Fig.~\ref{fig:Leyre_tests}, bottom-right shows the result of simply
setting these weights to $\frac{1}{4}$ and $S^{opt}$ to the average
values, as described in Section~\ref{subsec:Sopt_W}, while maintaining
$\bar{q}$ ratio sharpening. As expected, details are also smoothed
out. Weights $W$ are defined between high-resolution pixels, hence
encode high-resolution details. The reverse weights $V$ encode larger
range patterns present in surrounding low-resolution pixels. Results
presented in this section use the 60m/pixel band 1, for which two
super-resolution steps are applied. This choice was made so as to
enhance and clearly highlight the influence of shared values $S$,
of $\bar{q}$ ratio sharpening, and of weights $W$, $V$, on the
final result. For 20m/pixel bands only one super-resolution step is
applied, but all parts of the method are still needed for good results.

\section*{Conclusion}

This article presents a super-resolution method based on exploiting
both the local consistency between neighborhood pixels and the geometric
consistency of sub-pixel constituents across multispectral bands.
Figs.~\ref{fig:Coastal_area_RGB_B8_B1_B5}-\ref{fig:Fields_area_B9_B11_B12}
show the result of applying this method to Sentinel-2 images, in order
to bring all bands from 20m/pixel and 60m/pixel down to the highest
resolution at 10m/pixel. The algorithm is however generic and could
be applied to other multi-resolution and multispectral satellite images.
Further work could include the usage of secondary images, taken from
a satellite with low temporal resolution but with a higher spatial
resolution, in order to extract the geometric information used for
the super-resolution. Assuming the pixel geometry does not change
much between these acquisitions, then Sentinel-2 images could be enhanced
below 10m/pixel. This form of multi-satellite temporal super-resolution
would combine high temporal frequency with high spatial resolution.
Another trail of research would be to incorporate the super-resolution
algorithm directly within the atmospheric correction step \cite{Sen2Cor221},
rather than applying it as a separate stage. Indeed, using higher-resolution
pixels instead of low-resolution bands for calibrating the atmospheric
correction may lead to better accuracy.

\section*{Acknowledgments}

The author thanks Hussein Yahia and Dharmendra Singh for useful discussions
and feedback while carrying this work.

\section*{Source code}

The source code for this super-resolution algorithm is provided as
Free/Libre software, under either (your choice) the lesser GNU public
licence v2.1 or more recent, or the CeCILL-C licence. The library
is written and usable directly in C++ and it is also wrapped in a
Python package. A ready to use Python script for super-resolving Sentinel-2
images is provided. See \url{http://nicolas.brodu.net/recherche/superres/}.

\section*{Appendix : Sentinel 2 bands}

ESA's Sentinel-2 program comprises two satellites (2A and 2B) with
identical characteristics:
\begin{center}
\begin{tabular}{|c||c|c|c|}
\hline 
Band & Central wavelength & Bandwidth & Pixel size\tabularnewline
\hline 
\hline 
B2 & 490 nm & 65 nm & 10 m\tabularnewline
\hline 
B3 & 560 nm & 35 nm & 10 m\tabularnewline
\hline 
B4 & 665 nm & 30 nm & 10 m\tabularnewline
\hline 
B8 & 842 nm & 115 nm & 10 m\tabularnewline
\hline 
\hline 
B5 & 705 nm & 15 nm & 20 m\tabularnewline
\hline 
B6 & 740 nm & 15 nm & 20 m\tabularnewline
\hline 
B7 & 783 nm & 20 nm & 20 m\tabularnewline
\hline 
B8A & 865 nm & 20 nm & 20 m\tabularnewline
\hline 
B11 & 1610 nm & 90 nm & 20 m\tabularnewline
\hline 
B12 & 2190 nm & 180 nm & 20 m\tabularnewline
\hline 
\hline 
B1 & 443 nm & 20 nm & 60 m\tabularnewline
\hline 
B9 & 945 nm & 20 nm & 60 m\tabularnewline
\hline 
B10 & 1375 nm & 30 nm & 60 m\tabularnewline
\hline 
\end{tabular}
\par\end{center}

Band 10 is dedicated for cloud detection and removed by the bottom-of-atmosphere
correction utility \cite{Sen2Cor221}. The algorithm presented in
this paper nevertheless takes it into account when applied to the
unprecessed, top-of-atmosphere images.

\end{document}